\documentclass[fleqn,10pt]{wlpeerj}
\usepackage{bbm}
\title{Analysis and tuning of hierarchical topic models based on Renyi entropy approach}

\author[1]{Sergei Koltcov}
\author[1]{Vera Ignatenko}
\author[1]{Maxim Terpilovskii}
\author[2, 1]{Paolo Rosso}
\affil[1]{Laboratory for Social and Cognitive Informatics, National Research University Higher School of Economics, St. Petersburg, Russia}
\affil[2]{PRHLT Research Center, Universitat Politècnica de València, Valencia, Spain}
\corrauthor[1]{Sergei Koltcov}{skoltsov@hse.ru}

\begin{abstract}
Hierarchical topic modeling is a potentially powerful instrument for determining the topical structure of text collections that allows constructing a topical hierarchy representing levels of topical abstraction. However, tuning of parameters of hierarchical models, including the number of topics on each hierarchical level, remains a challenging task and an open issue. In this paper, we propose a Renyi entropy-based approach for a partial solution to the above problem. First, we propose a Renyi entropy-based metric of quality for hierarchical models. Second, we propose a practical concept of hierarchical topic model tuning tested on datasets with human mark-up. In the numerical experiments, we consider three different hierarchical models, namely, hierarchical latent Dirichlet allocation (hLDA) model, hierarchical Pachinko allocation model (hPAM), and hierarchical additive regularization of topic models (hARTM). We demonstrate that hLDA model possesses a significant level of instability and, moreover, the derived numbers of topics are far away from the true numbers for labeled datasets. For hPAM model, the Renyi entropy approach allows us to determine only one level of the data structure. For hARTM model, the proposed approach allows us to estimate the number of topics for two hierarchical levels. 
\end{abstract}

\begin{document}

\flushbottom
\maketitle
\thispagestyle{empty}

\section*{Introduction}
The vast flow of information generated by various TV news channels, electronic newspapers, and magazines is most often represented as a hierarchical system. In such a system, news items are divided into a number of global topics, such as politics, sports, covid19, and others. Within each of the main topics, the documents reflect a news diversity on a given topic. The hierarchical division of information content is highly convenient and seems to reflect specific cognitive characteristics of a person \citep{Cohen:2006, Palmer:1977, Taylor:2015}. Therefore, starting from 2004, the active development of probabilistic topic models that allow for identifying the topical hierarchical structure in large datasets has begun \citep{Blei:2003h, Li:2006}. However, each of these models has a set of parameters, which have to be tuned to  obtain a topical solution of higher quality. Correspondingly, a problem of tuning hierarchical topic models arises. A solution to this problem is complicated by a set of factors. First, there are no generally accepted and appropriate  metrics of quality that take into account the features of hierarchical modeling. Second, the amount of publicly available datasets with the hierarchical mark-up, which can be used to tune and compare hierarchical models, is very limited. Third, when applying hierarchical topic models on real datasets, the type of topical structure  (non-hierarchical or hierarchical, and also the number of levels in the case of hierarchical) of data is unknown in advance. Forth, hierarchical topic models, as well as flat topic models, possess a certain level of semantic instability, which means that different runs of the algorithm on the same source data lead to different solutions. This complicates the search for optimal model parameters on a given dataset. Thus, investigation and assessment of the ability to tune hierarchical topic models is an urgent task. 

In this work, we investigate the behavior of three hierarchical models (hierarchical latent Dirichlet allocation (hLDA) \citep{Blei:2003h}, hierarchical Pachinko allocation (hPAM) \citep{Mimno:2007}, and hierarchical additive regularization of topic models (hARTM) \citep{Chirkova:2016}) in terms of two metrics: log-likelihood and Renyi entropy. We conduct experiments on four marked-up collections, two of which are non-hierarchical and two others have two-level hierarchical mark-up. 
The goal of our research is, first, to estimate the ability of hierarchical models to identify a hierarchical or non-hierarchical structure in the data and, second, to find the best metric of quality that is suitable for tuning of hierarchical models.  In this research framework, we propose an extension of an entropic approach, which was developed earlier for non-hierarchical topic models, for a partial solution to the above problems. First, we propose a Renyi entropy-based metric of quality for hierarchical models. Second, we propose a practical concept of hierarchical topic model tuning tested on datasets with human mark-up.

To simplify the structure of this work, an overview of hierarchical models and existing metrics of quality is provided in Appendix A. Thus, our work consists of the following parts. Section 'Entropic approach to hierarchical topic modeling' contains the description of the entropic approach for estimating the quality of topics models. Section 'Description of computer experiments' reviews the features of our computer experiments for each of the considered models. Section 'Numerical results' contains an analysis of the behavior of hierarchical models under variation of hyperparameters and the number of topics for the three models on four datasets. Section 'Discussion' describes the obtained results and reviews the possible limitations of the proposed approach. 
Section 'Conclusion' summarizes our findings and
and contains practical recommendations for choosing a hierarchical model.
Appendices B and C contain tables summarizing the numerical results for hPAM  and hLDA models, correspondingly.





\section*{Entropic approach to hierarchical topic modeling}
The entropic approach for topic models is based on the concept that an information system consisting of a large number of words and documents represents a statistical system. The state of such a statistical system can be characterized by a value of entropy. It is well known that the maximum entropy of a statistical system corresponds to either chaos or a uniform distribution of the system elements. However, unlike real physical systems, a text collection can be subject to procedures (for instance, clustering or topic modeling) that change the value of entropy by ordering the data. 
Based on the fact that entropy equals minus information $S=-I$ \citep{Beck:2009}, one can implement the process of document clustering or topic modeling in such a way that leads to entropy minimum (information maximum), which corresponds to highly non-equilibrium distribution. 
Since the procedure of clustering of text collections significantly depends on the values of model parameters, which include the parameter 'number of topics', an algorithm of model tuning can be organized in the form of searching for parameters that would lead to minimum entropy. Let us note that modern widely used topic models usually do not include hyperparameters optimization algorithms (or include only for some of them). Thus, this is the user who has to select the values of model parameters based on some external metrics of quality.

Recent research \citep{Koltsov:2018, Koltcov:2020} demonstrated that the most convenient metric for estimating the state of a textual statistical system is Renyi entropy, whose calculation is entirely based on the obtained in the process of topic modeling probabilities of words to belong to a particular cluster or topic (matrix $\Phi=\{\phi_{wt}\}$, where $\phi_{wt}$ is the probability of word $w$ in topic $t$; for more details we refer the reader to Appendix A). Calculation of Renyi entropy for a topic model is based on two observable variables: 1) Gibbs-Shannon entropy of the model; 2) internal energy of the model. Gibbs-Shannon entropy $S$ can be calculated as follows: $S= \ln(\rho)=\ln(\frac{N}{WT})$, where $N$ is the number of words with $\phi_{wt}>1/W$, $W$ is the number of unique words in the dataset, $T$ is the number of topics. The internal energy $E$ of a topic model can be expressed in the following way: $E =-\ln(\tilde{P})=-\ln \left(\frac{1}{T} \sum_{wt}(\phi_{wt}\cdot \mathbbm{1}_{\{\phi_{wt}>1/W\}})\right)$, where $\sum_{wt}(\phi_{wt}\cdot \mathbbm{1}_{\{\phi_{wt}>1/W\}})$ is the sum of probabilities of words, each of which is above the threshold $1/W$. Then, Renyi entropy can be calculated through free energy ($F=-qE+S$) in the following way:

\begin{equation}
     S_q^R = \frac{F}{q-1}= \frac{q\ln(q \tilde{P})+q^{-1}\ln(\rho)}{q-1},
\end{equation}

where the deformation parameter $q=1/T$ is the inverse number of topics. Thus, Renyi entropy of a topic model explicitly contains the parameter 'number of topics' in the form of a deformation parameter. Application of Renyi entropy for tuning different non-hierarchical topic models is considered in work \citep{Koltcov:2020}.

As shown in works \citep{Blei:2003h, Mimno:2007}, hierarchical structure in topic models can be represented as a graph, where each node on a chosen level represents a topic of the corresponding hierarchical level of topic abstractions. Each topic $t$ is characterized by the words with corresponding probabilities $\phi_{wt}$ of belonging to this topic. Thus, the hierarchical topic modeling procedure leads to the construction of the hierarchical tree with a fixed number of topics on each level. The total number of words on each level is constant and equals the total number of elements, $W$, in the statistical system. The set of nodes-topics at one level of the hierarchy represent matrix $\Phi$ (distribution of words by topics). Thus, the hierarchical modeling procedure leads to the construction of a sequence of matrices $\Phi^i$ ($i$ refers to the level $i$ in the hierarchy), where the number of words is constant, and the number of topics is subsequently increased. The ratio of words with high probabilities (i.e., probabilities above the threshold) changes when shifting from a level to the next level.  Therefore, each level of the hierarchy can be characterized by the following variables: 1) the number of topics $T_i$ on level $i$; 2) the number of words ($N_i$) with high probabilities: $\phi_{wt}^{i}>1/W$; 3) the sum of probabilities above the threshold:
$\tilde{P}_i=\sum_{wt}\phi_{wt}^{i}\cdot \mathbbm{1}_{\{\phi_{wt}^{i}>1/W\}}$. Based on these variables, one can calculate the internal energy $E_i$ and Gibbs-Shannon entropy $S_i$ of the current level $i$ with respect to the equilibrium state of that level: $E_i=-\ln(\frac{\tilde{P}_i}{T_i})$, $S_i=\ln(\frac{N_i}{WT_i})$. Using $S_i$ and $E_i$, one can determine free energy and Renyi entropy of level $i$. Free energy of hierarchical level $i$ is expressed as follows: $F_i=E_i-T_i \cdot S_i$. Renyi entropy of level $i$ can be expressed in the following way: $S_i^R=\frac{F_i}{1-q}$, where $q=1/T_i$ is the deformation parameter characterizing each level of the hierarchy.  

Thus, by measuring the value of entropy on each level and varying number of topics and other hyperparameters for a particular dataset, one can estimate the process of hierarchy construction from the point of view of the behavior of $S_i^R$ when moving from level to level in the hierarchy, i.e. from the point of view of information theory. Here, the process of clustering of words by topics starts with the minimum information (maximum entropy) when all the elements (words) of the statistical system are assigned to one or two topics and ends also with the maximum entropy when all the elements are almost uniformly distributed among topics (when the number of topics is large). The locations of the global minimum and a number of local minima of Renyi entropy, as a function of the number of topics, is determined by the features of the data. Renyi entropy $S_i^R$ serves as a measure of the degree to which a given system is non-equilibrium, where minimum entropy corresponds to information maximum.

\section*{Results}
\subsection*{Data}
In our numerical experiments, the following datasets were used:
\begin{itemize}
  \item 'Lenta' dataset (from lenta.ru news agency, available at https://www.kaggle.com/yutkin/corpus-of-russian-news-articles-from-lenta). This dataset  contains 8,630 documents with a vocabulary of 23,297 unique words in the Russian language. Each of these documents is manually assigned with a class from a set of 10 topic classes. Since some of the topics are strongly correlated, the documents in this dataset can be represented by 7-10 topics.
  \item '20 Newsgroups' dataset (http://qwone.com/~jason/20Newsgroups/) is a widely used dataset in the field of topic modeling. This dataset consists of  15,425 news articles with 50,965 unique words. Each of the news items is assigned to one of 20 topic groups. Since some of these topics can be combined, 14-20 topics can represent this dataset's documents according to \citep{Basu}.
  \item 'WoS' dataset (available at https://data.mendeley.com/datasets/9rw3vkcfy4/1) is a dataset with hierarchical two-level mark-up. The original dataset  contains 46,985  abstracts of published papers available from the \textit{Web of Science}. The first level contains 7 categories (domains): computer science, electrical engineering, psychology, mechanical engineering, civil engineering, medical science, and biochemistry. The seconds level contains 134 specific topics (areas), each of which belongs to one of the categories. The number of unique words is 80,337. This dataset is often used as a benchmark dataset for hierarchical classification \citep{Sinha:2018}.
However, this dataset is highly unbalanced concerning the distribution of the number of documents per sub-categories. For instance, some sub-categories contain more than 700 documents and some contain less than 50 documents. Therefore, we also consider a balanced subset of this dataset (described below), where poorly presented topics, i.e., topics with a small number of documents, were deleted.  
\item 'Balanced WoS' dataset (available at https://data.mendeley.com/datasets/9rw3vkcfy4/1) is a class-balanced subset of the 'WoS' dataset, which contains 11,967 abstracts. The first level contains 7 categories and the seconds level contains 33 areas.
\item 'Amazon' dataset (available at https://www.kaggle.com/kashnitsky/hierarchical-text-classification/
version/1) is a dataset with hierarchical three-level mark-up. It contains 40,000 product reviews from \textit{Amazon}. The vocabulary of this dataset consists of 31,486 unique words. Level 1 of the hierarchical mark-up contains 6 categories, level 2 contains 64 child categories, and level 3 contains 510 categories. We consider only the first two levels of the provided hierarchical classes since the third level contains 'unknown' labels.  Let  us note that the original dataset is highly imbalanced. Some sub-categories contain less than 50 documents, while some other sub-categories contain more than 2,000 documents. Therefore, a balanced subset of this dataset is also considered.
\item 'Balanced Amazon' dataset is a subset of the 'Amazon' dataset that contains only sub-categories with the number of documents above 500. So, level 1 contains 6 categories and level 2 contains 27 sub-categories. The total number of documents is 32,774, and the number of unique words is 28,422. 
\end{itemize}

\subsection*{Description of computer experiments}
 In our numerical experiments, topic modeling was performed on the above four datasets (two datasets with non-hierarchical mark-up and two datasets with two-level mark-up) for each model. Correspondingly, the results are divided into two parts. In the first part, we analyze the applicability of the models for the datasets with a non-hierarchical structure. In the second part, we consider the behavior of the models on the datasets with hierarchical mark-up. Since topic modeling possesses a certain level of instability that leads to fluctuations in the word probabilities, all calculations were performed at least six times (for each combination of hyperparameters), and then the results were averaged. The features of experiments for each model are discussed below.

\subsubsection*{hPAM model}
Let us note that hPAM model depends on the following parameters: 1) The number of topics on the second and the third level; 2) hyperparameter $\eta$; 3) hyperparameter $\alpha$. The number of topics on the first level of hPAM model is always set to one. Moreover, in this model, the user can only set the initial value of parameter $\alpha$, and then the algorithm tunes it during the modeling. For a more detailed description of the model, we refer the reader to Appendix A.
Our computer experiments demonstrated that variation of the initial value of $\alpha$ does not influence the results of modeling. Therefore, in the rest of the paper, we do not vary this value. 
In our work, hPAM model was studied in two stages. 
At the first stage, the number of topics on the third level was fixed $T_2=1$ while hyperparameter $\eta$ was varied in the range [0.001, 1] and the number of topics on the second level ($T_1$) was varied in the range [2, 200]. For each solution, we calculated Renyi entropy of the second level of the hierarchy and log-likelihood. Then, the values of $T_1$ and $\eta$ that corresponded to minimum Renyi entropy were found and fixed. At the second stage, the number of topics on the third level was varied under the condition of fixed $T_1$ and $\eta$. For each combination of parameters, Renyi entropy of the third level was calculated.

\subsubsection*{hLDA model} 
Let us note that hLDA model has the following parameters: 1) depth of the hierarchy; 2) hyperparameter $\alpha$, which is tuned by the model automatically; 3) hyperparameter $\gamma$; 4) hyperparameter $\eta$. hLDA model is non-parametric; therefore, it infers the number of topics on each level automatically.  In this work, we studied the dependence  of the number of inferred topics on the parameter $\eta$, which was varied in the range $[0.001,1]$. Since this model is highly unstable and can produce different numbers of topics for runs with the same values of hyperparameters, we ran the model 10 times for each value of $\eta$. Then, we estimated the range of the derived number of topics on the second and the third levels. The mean values of Renyi entropy and log-likelihood were calculated as well.  

\subsubsection*{hARTM model} 
hARTM model has the following parameters: 1) the number of topics on each level of the hierarchy; 2) seed is a parameter describing the initialization procedure (it sets a random number generator). This model was also studied in two stages. At the first stage, the number of topics on the second level was fixed ($T_2=1$), and the number of topics on the first level was varied in the range of [2, 200] topics. Based on the minimum Renyi entropy location, the optimal $T_1$ was chosen. At the second stage, the number of topics on the second level was varied under the condition of fixed $T_1$.   

\subsection*{Numerical results}
\subsubsection*{hPAM model (the first stage)}
\begin{figure}
\centering
\includegraphics[width=3.3in]{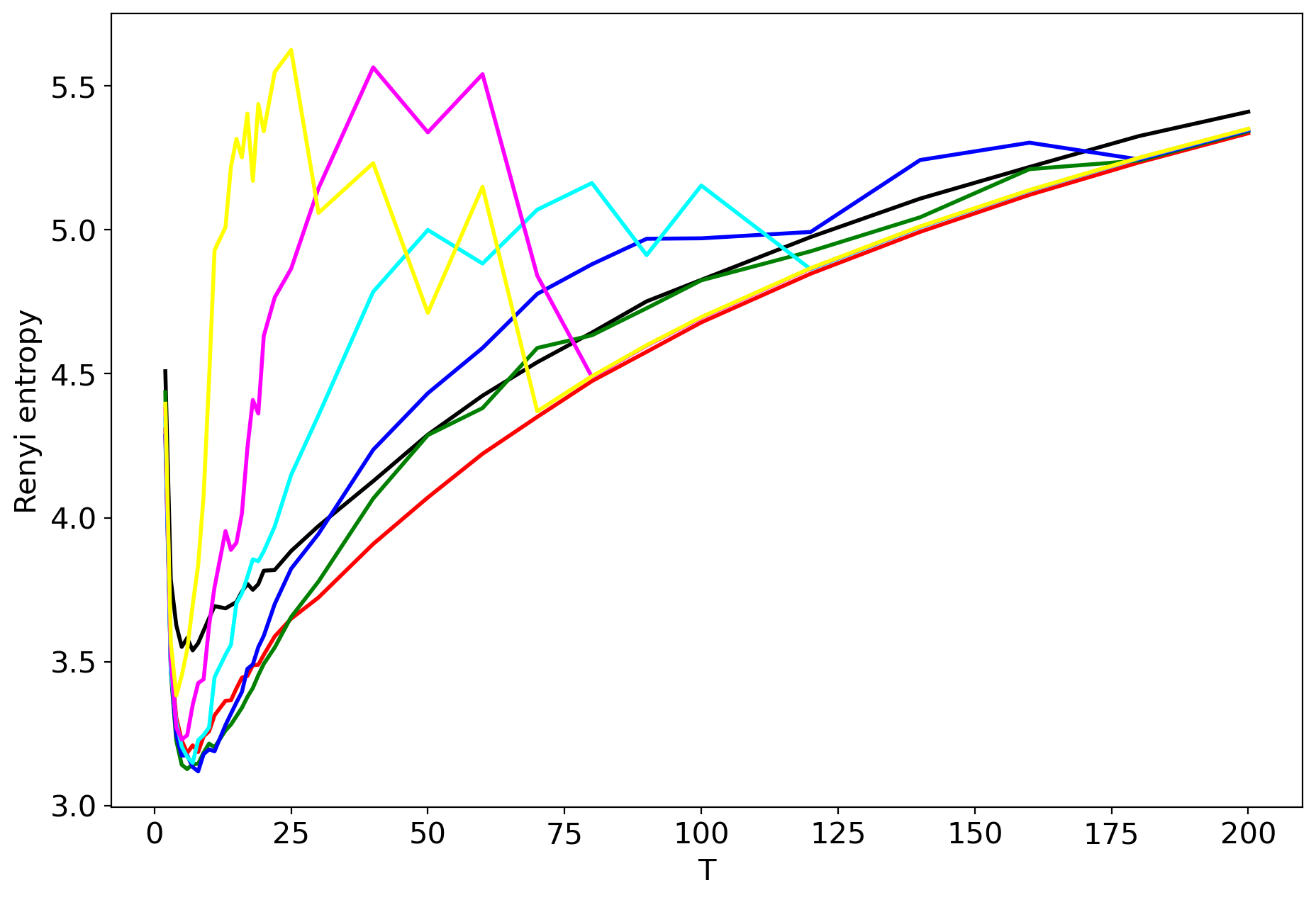}
\caption{Renyi entropy curves (hPAM). Black: $\eta=0.001$; red: $\eta=0.01$; green: $\eta=0.2$; blue: $\eta=0.3$; cyan: $\eta=0.5$; magenta: $\eta=0.7$; yellow: $\eta=1$. Lenta dataset.}
\label{fig:HPAM_lenta_Renyi_lev2}
\end{figure}

\begin{figure}
\centering
\includegraphics[width=3.3in]{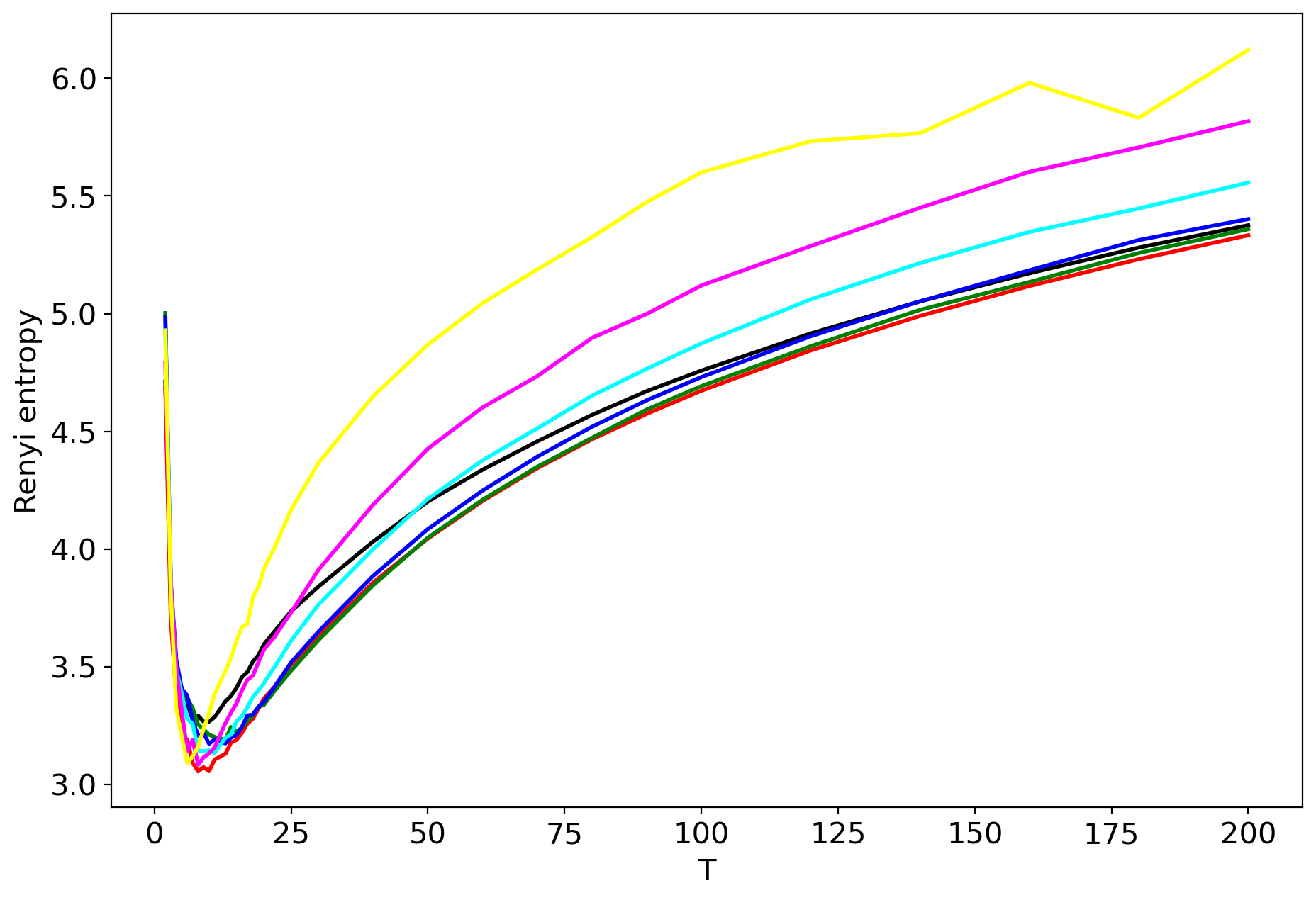}
\caption{Renyi entropy curves (hPAM). Black: $\eta=0.001$; red: $\eta=0.01$; green: $\eta=0.2$; blue: $\eta=0.3$; cyan: $\eta=0.5$; magenta: $\eta=0.7$; yellow: $\eta=1$. 20 Newsgroups dataset.}
\label{fig:HPAM_20news_Renyi_lev2}
\end{figure}

\begin{figure}
\centering
\includegraphics[width=3.3in]{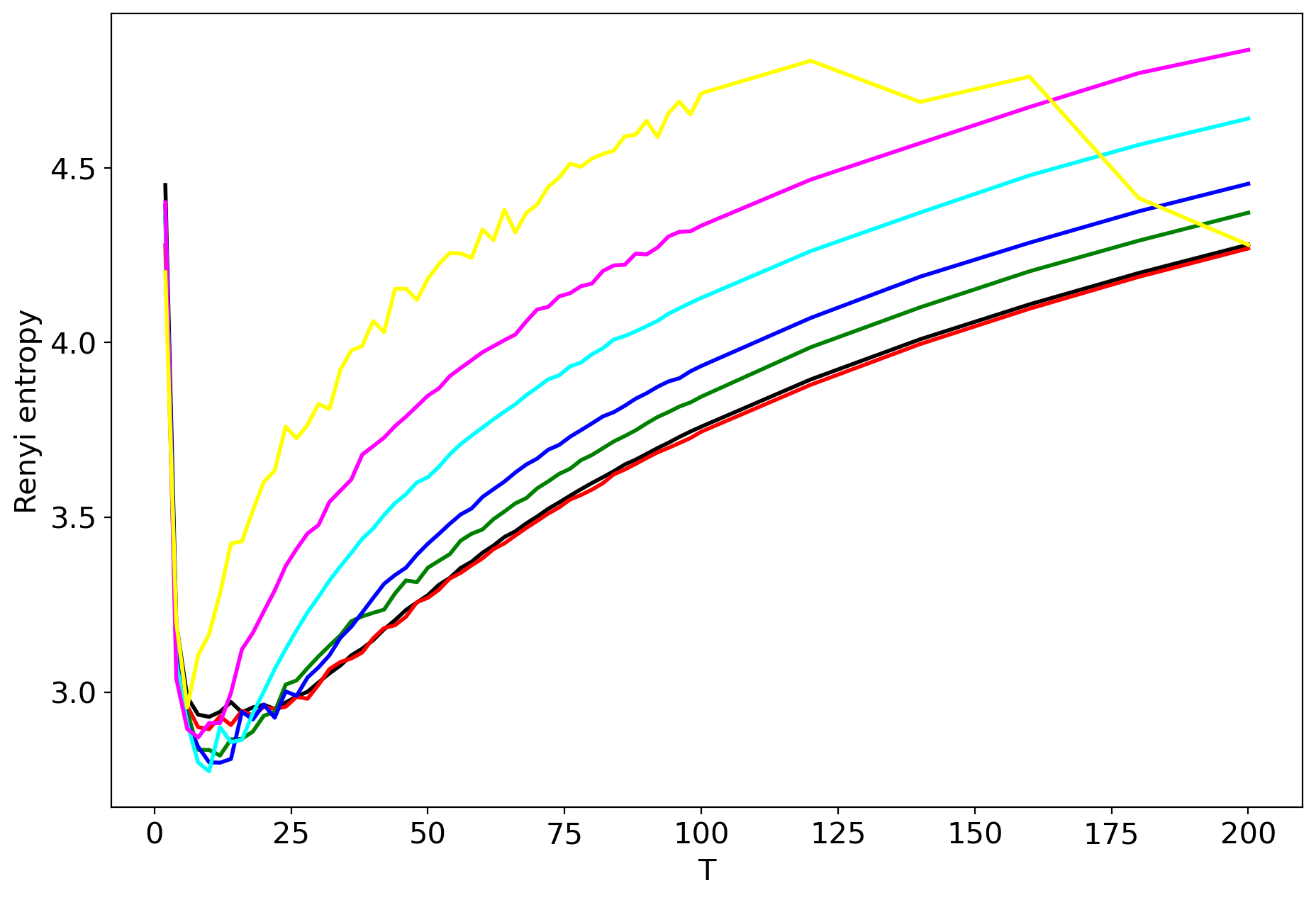}
\caption{Renyi entropy curves (hPAM). Black: $\eta=0.001$; red: $\eta=0.01$; green: $\eta=0.2$; blue: $\eta=0.3$; cyan: $\eta=0.5$; magenta: $\eta=0.7$; yellow: $\eta=1$. Balanced WoS dataset.}
\label{fig:HPAM_WoSbalanced_Renyi_lev2}
\end{figure}

\begin{figure}
\centering
\includegraphics[width=3.3in]{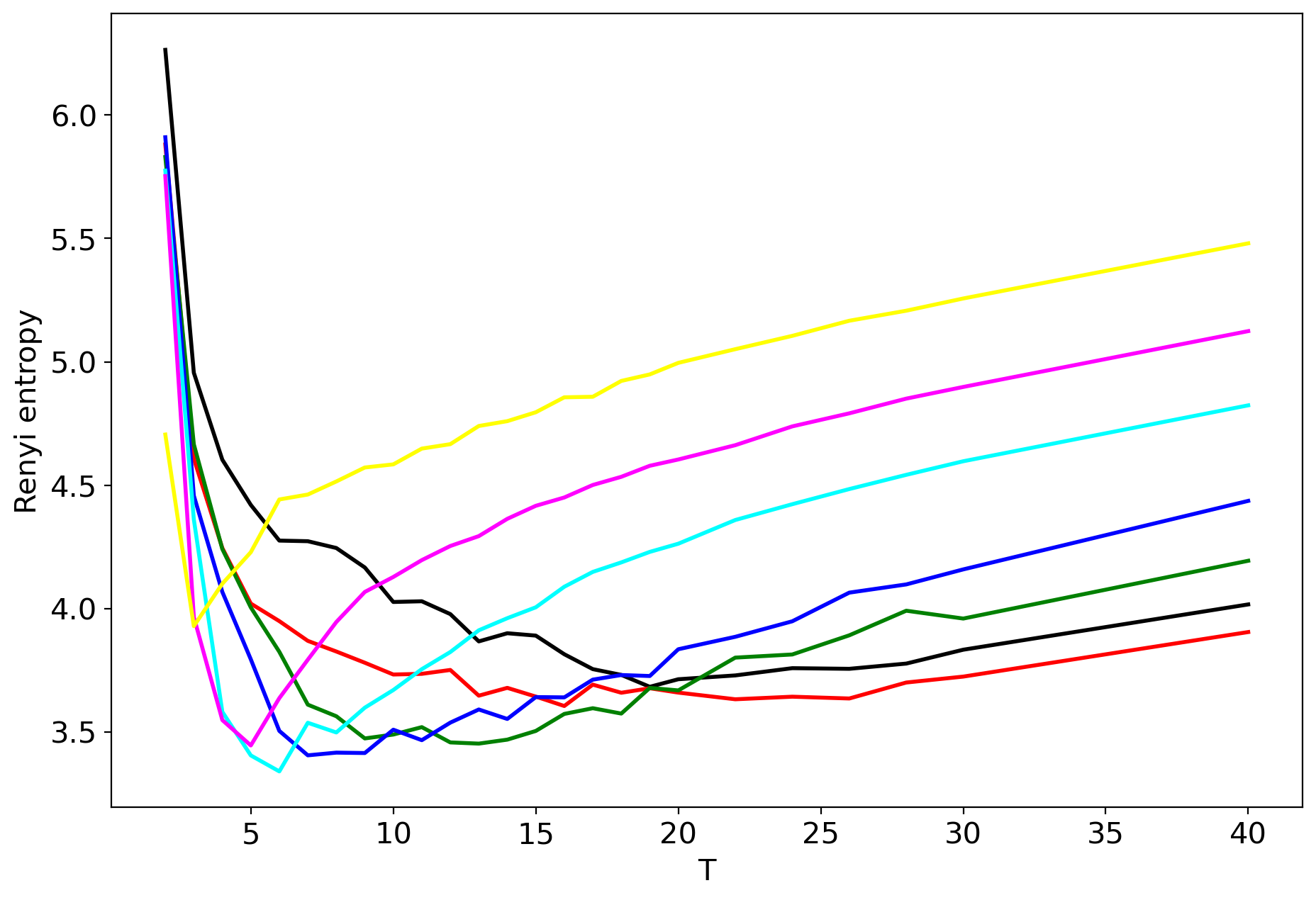}
\caption{Renyi entropy curves (hPAM). Black: $\eta=0.001$; red: $\eta=0.01$; green: $\eta=0.2$; blue: $\eta=0.3$; cyan: $\eta=0.5$; magenta: $\eta=0.7$; yellow: $\eta=1$. Balanced Amazon dataset.}
\label{fig:HPAM_Amazonbalanced_Renyi_lev2}
\end{figure}

Figures \ref{fig:HPAM_lenta_Renyi_lev2}-\ref{fig:HPAM_Amazonbalanced_Renyi_lev2} demonstrate the behavior of Renyi entropy under variation of $\eta$ and the number of topics on the second level for different datasets. The common pattern of Renyi entropy behavior for all the datasets occurs for small (about 2-3) and large (about 100-200) numbers of topics that corresponds to two extreme states of the statistical system characterized by entropy maximum. Moreover, one can see that the location and the value of minimum Renyi entropy significantly depend on the parameter $\eta$. Large values of $\eta$ ($\eta>0.7$) lead to large fluctuations in the Renyi entropy for large numbers of topics that complicates finding entropy minimum when the number of topics increases. Correspondingly, the further increment of $\eta$ is inadvisable. The behavior of Renyi entropy for this model allows us to find approximations of the optimal number of topics on the second level for different datasets  and the optimal value of $\eta$ by means of selecting the values that correpond to the minimum entropy. Since we test the model on the datasets with human mark-up, we can estimate the error of the found approximation of the number of topics. For non-hierarchical datasets (Lenta and 20 Newsgroups), the error corresponds to $\pm 2$ topics. For hierarchical datasets we obtain the error of $\pm 4$ topics (figure \ref{fig:HPAM_Amazonbalanced_Renyi_lev2}). The list of values of $\eta$ and corresponding numbers of topics with minimum Renyi entropy is given in Appendix B.  

\begin{figure}
\centering
\includegraphics[width=3.3in]{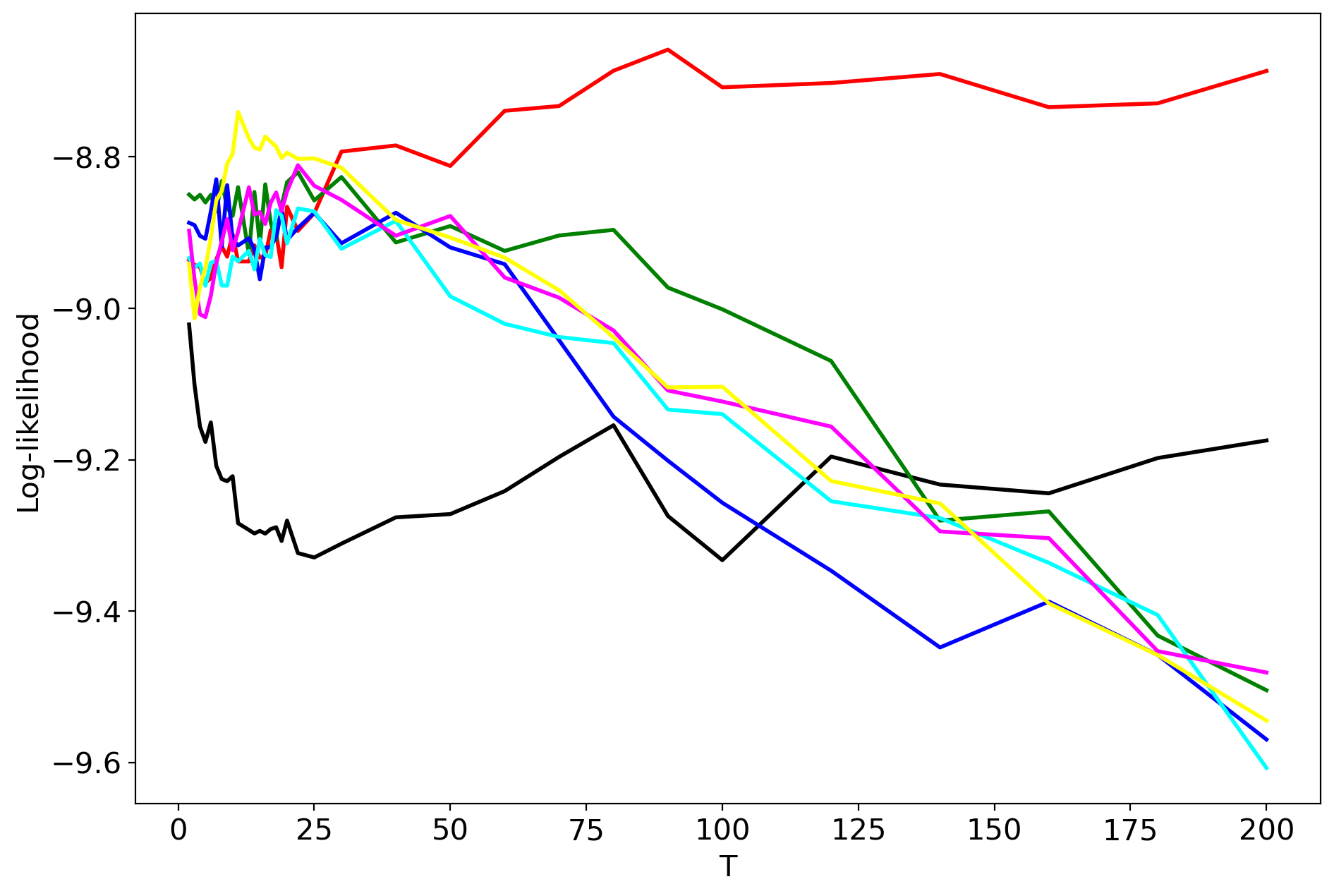}
\caption{Log-likelihood curves (hPAM). Black: $\eta=0.001$; red: $\eta=0.01$; green: $\eta=0.2$; blue: $\eta=0.3$; cyan: $\eta=0.5$; magenta: $\eta=0.7$; yellow: $\eta=1$. Lenta dataset.}
\label{fig:HPAM_Lenta_likelihood_lev2}
\end{figure}

\begin{figure}
\centering
\includegraphics[width=3.3in]{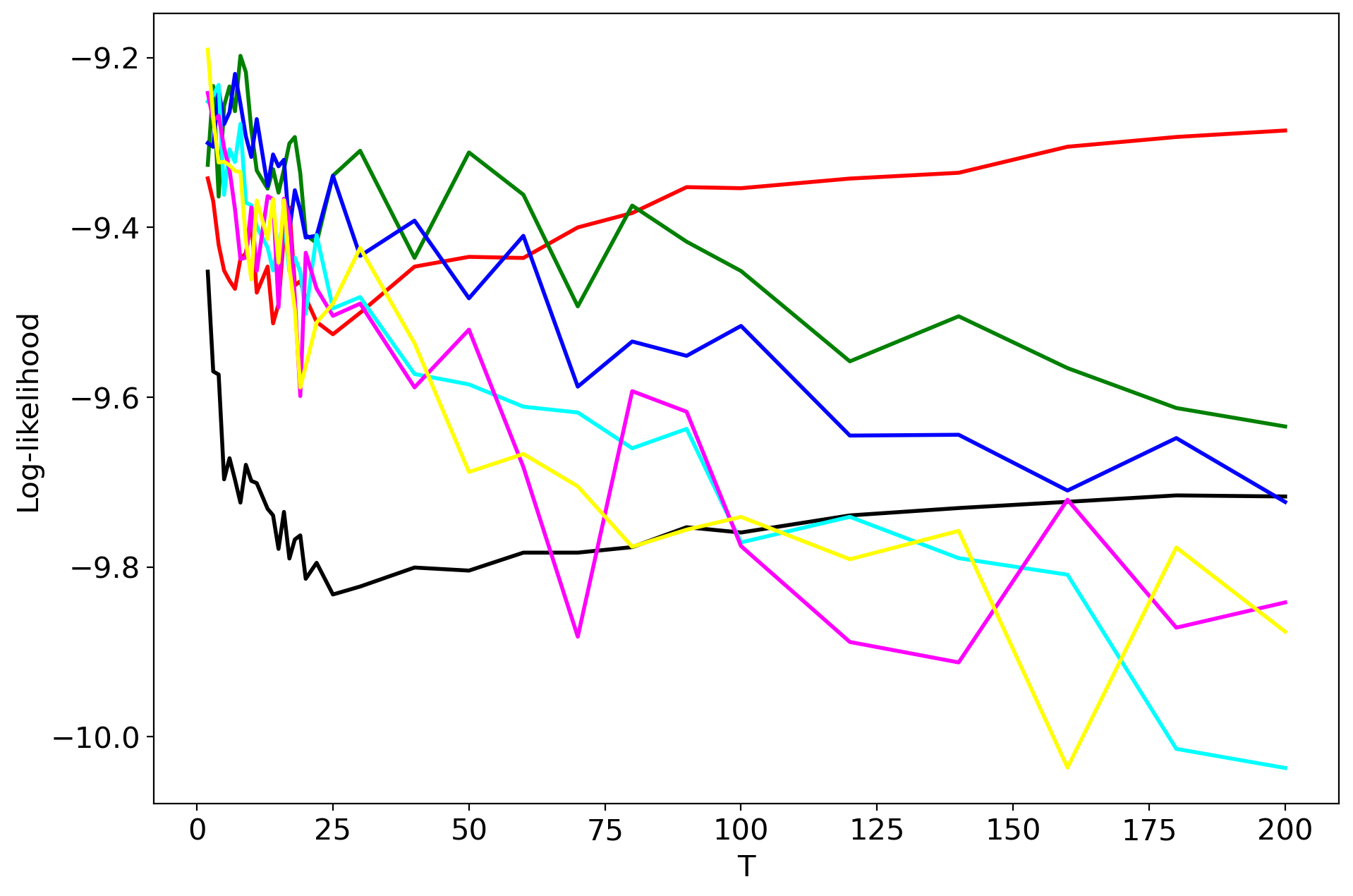}
\caption{Log-likelihood curves (hPAM). Black: $\eta=0.001$; red: $\eta=0.01$; green: $\eta=0.2$; blue: $\eta=0.3$; cyan: $\eta=0.5$; magenta: $\eta=0.7$; yellow: $\eta=1$. 20 Newsgroups dataset.}
\label{fig:HPAM_20news_likelihood_lev2}
\end{figure}

Calculation of log-likelihood under variation of $T_1$ and $\eta$ demonstrates that this metric is not useful for selecting the optimal values of $\eta$ and $T_1$ since it has large fluctuations (figures \ref{fig:HPAM_Lenta_likelihood_lev2} and \ref{fig:HPAM_20news_likelihood_lev2}).

\subsubsection*{hPAM model (the second stage)}
\begin{figure}
\centering
\includegraphics[width=3.3in]{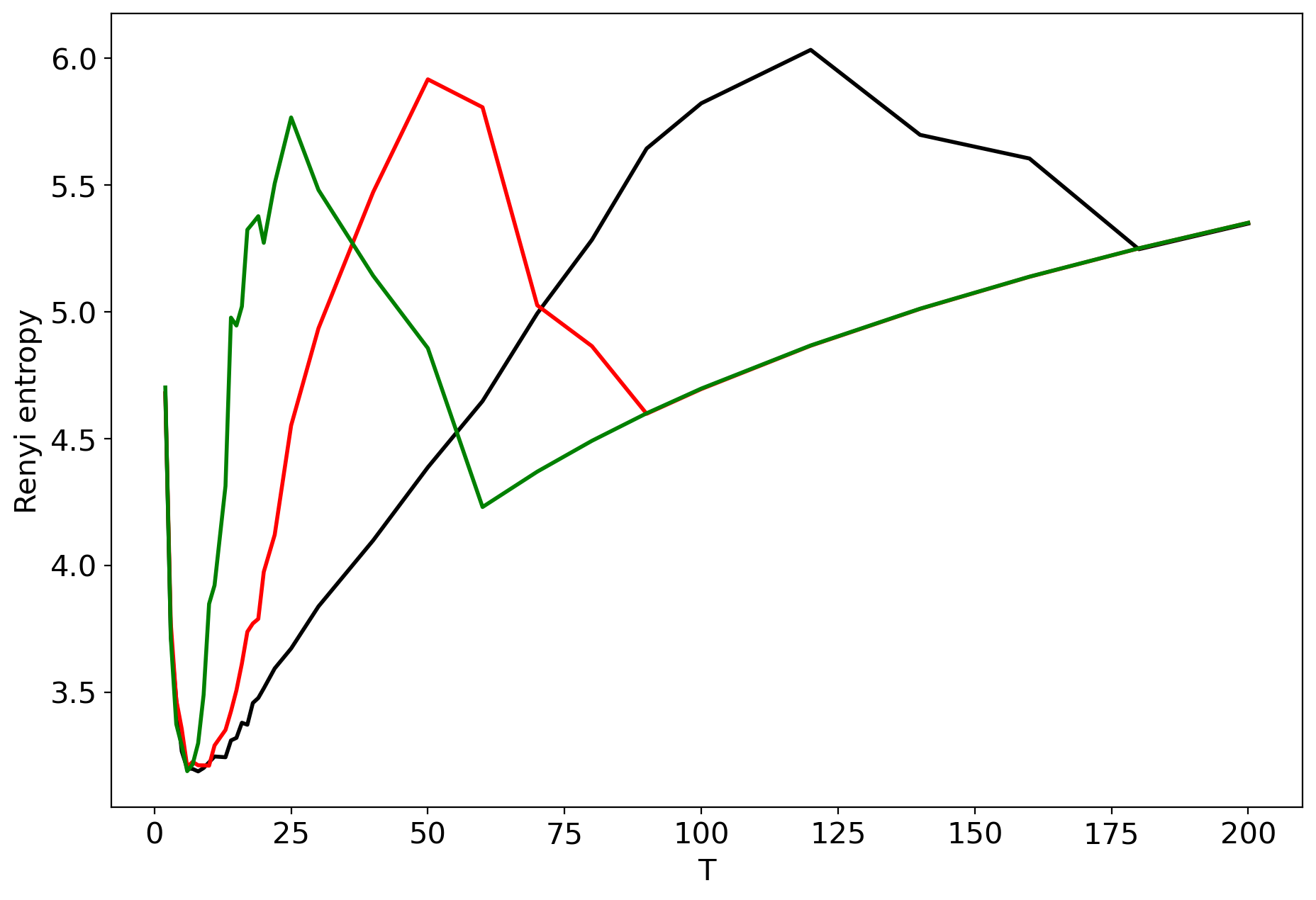}
\caption{Renyi entropy curves (hPAM). Black: $\eta=0.2$, $T_1=6$; red: $\eta=0.3$, $T_1=6$; green: $\eta=0.5$, $T_1=7$. Lenta dataset.}
\label{fig:HPAM_Lenta_Renyi_lev3}
\end{figure}

\begin{figure}
\centering
\includegraphics[width=3.3in]{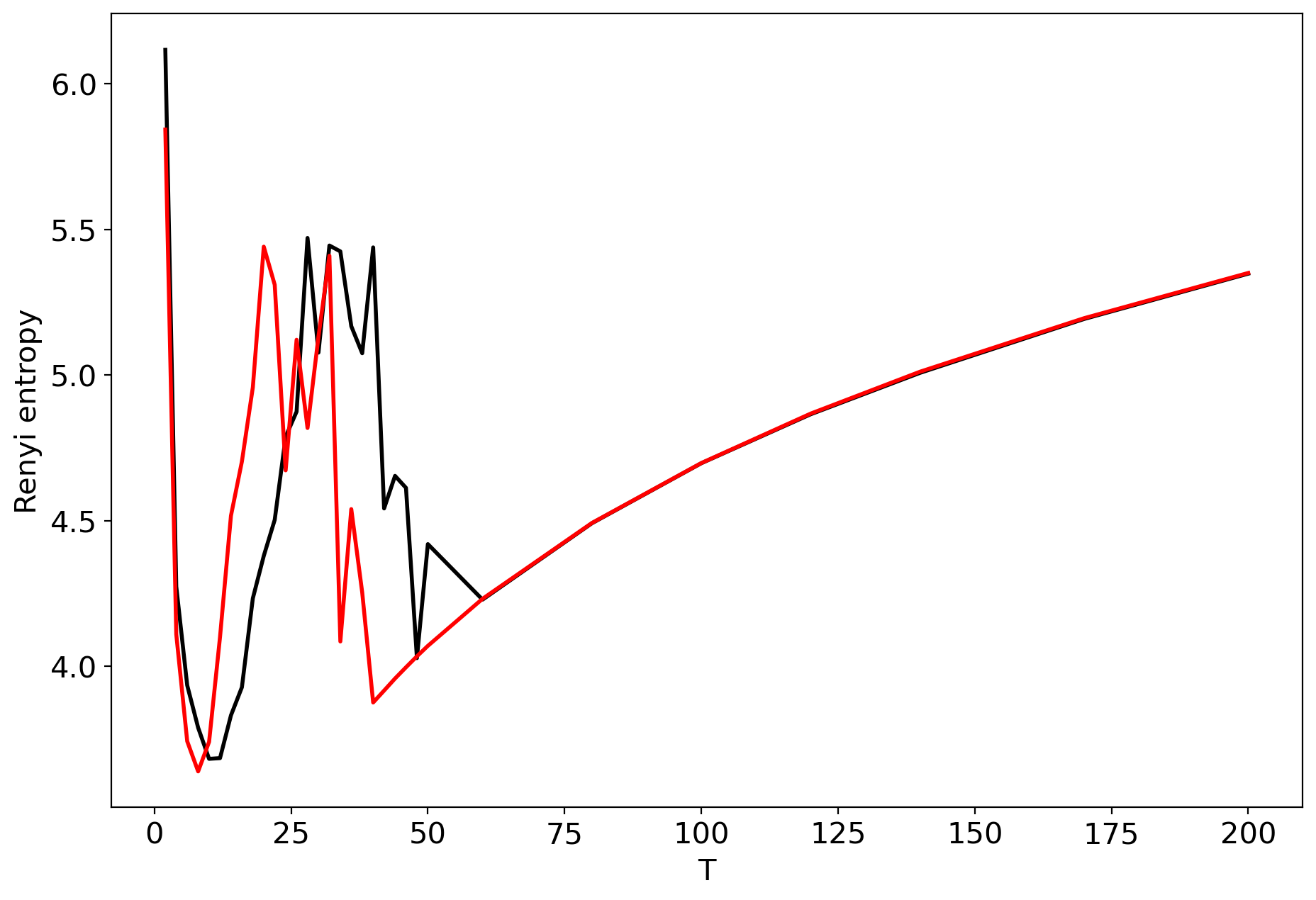}
\caption{Renyi entropy curves (hPAM). Black: $\eta=0.3$, $T_1=12$; red: $\eta=0.5$, $T_1=10$. Balanced WoS dataset.}
\label{fig:HPAM_WoSbalanced_Renyi_lev3}
\end{figure}

\begin{figure}
\centering
\includegraphics[width=3.3in]{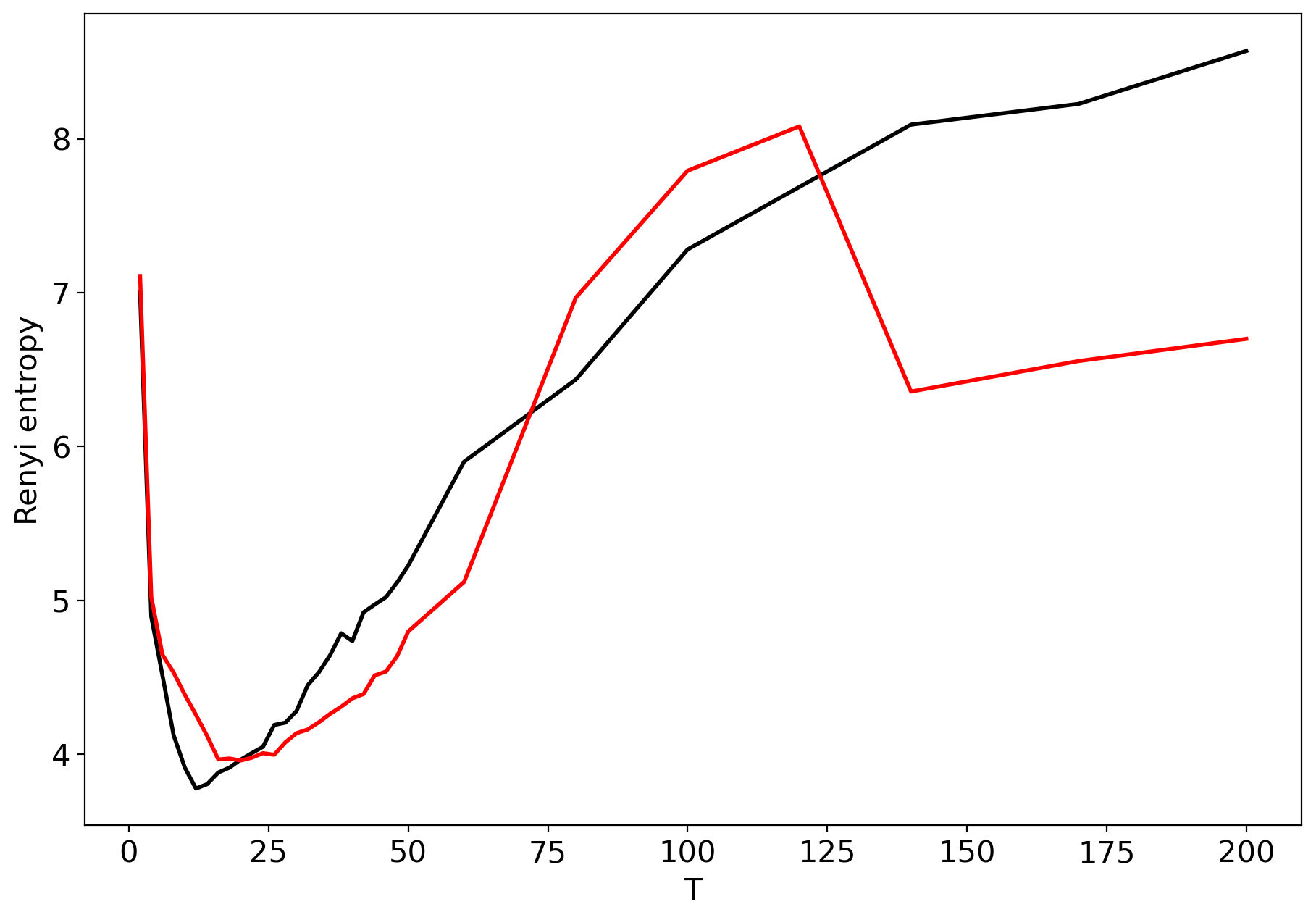}
\caption{Renyi entropy curves (hPAM). Black: $\eta=0.3$, $T_1=8$; red: $\eta=0.2$, $T_1=14$. WoS dataset.}
\label{fig:HPAM_WoS_Renyi_lev3}
\end{figure}

Figures \ref{fig:HPAM_Lenta_Renyi_lev3}-\ref{fig:HPAM_WoS_Renyi_lev3} demonstrate the behavior of Renyi entropy under variation of $T_2$ for chosen $T_1$ and $\eta$ from the first stage. For almost all the datasets except the 20 Newsgroups dataset, Renyi entropy curves have significant fluctuations and spikes. It should be noted that such fluctuations are typical for both balanced and unbalanced datasets. Moreover, parameter $\eta$ significantly influences the location of a spike. Therefore, the estimation of the number of topics on the third level is much less accurate than that of the second level. Our calculations demonstrate that in the region of large fluctuations, the model deteriorates. It leads to the fact that the number of words with high probabilities becomes constant and does not change with the increment of the number of topics. Correspondingly, the sum of high probabilities also becomes constant, i.e., statistical features of obtained solutions do not change. In this case, Renyi entropy changes only because the number of topics changes while other variables are constant. Thus, due to the features of hPAM model, the selection of the number of topics on the third level is complicated. 
Moreover, the log-likelihood metric does not allow us to choose the right number of topics for hPAM model nor determine whether the dataset has a hierarchical or non-hierarchical structure.

\subsubsection*{hLDA}
Our calculations demonstrate that hLDA model cannot be used in real applications since it infers very different numbers of topics for different runs with the same parameters. Moreover, the inferred numbers of topics are far away from the true number for considered datasets. In addition, the order of dispersion and the mean value of the predicted number of topics  on each level significantly depends on the value of parameter $\eta$. Due to high instability for $\eta<0.3$ and incorrect output for $\eta>0.3$, there is no sense in applying Renyi entropy approach to this model. Moreover, log-likelihood metric does not allow us to choose the right number of topics. The results of our calculations for hLDA model on different datasets are summarized in Appendix C. 

\subsubsection*{hARTM}
\begin{figure}
\centering
\includegraphics[width=3.3in]{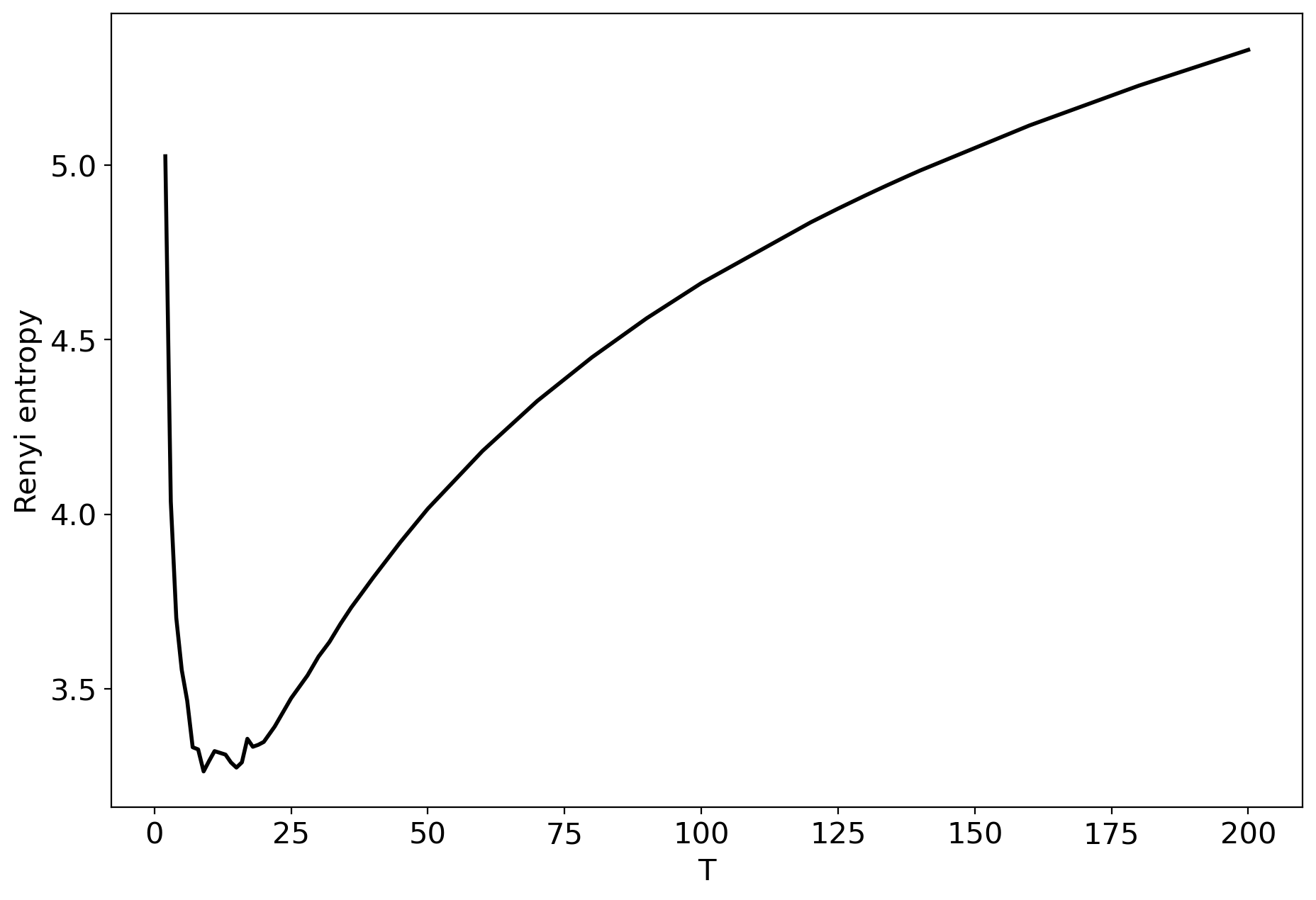}
\caption{Renyi entropy curve (hARTM). 20 Newsgroups dataset.}
\label{fig:HARTM_20news_Renyi_lev1}
\end{figure}

\begin{figure}
\centering
\includegraphics[width=3.3in]{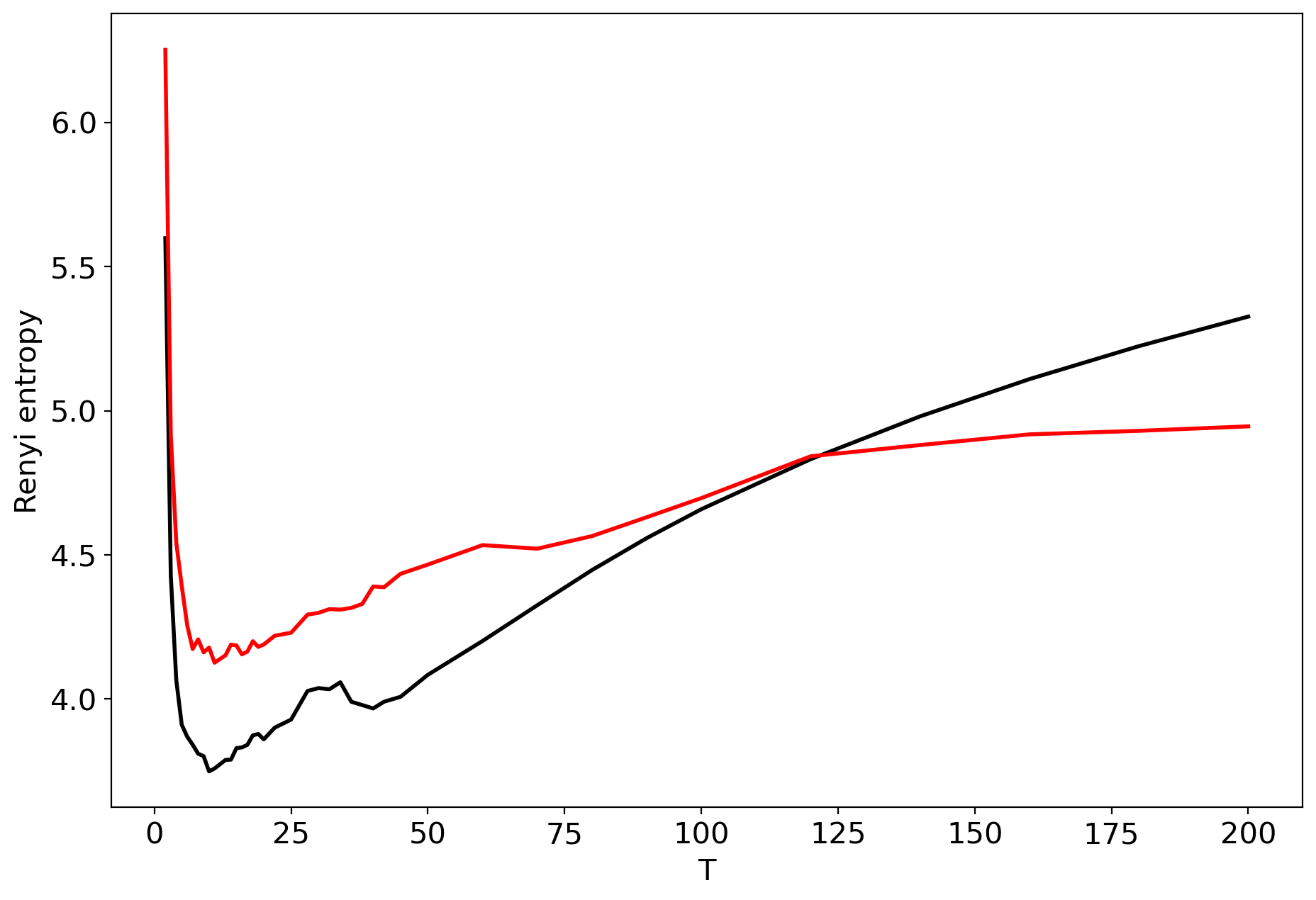}
\caption{Renyi entropy curves (hARTM). Black: balanced WoS dataset; red: WoS dataset.}
\label{fig:HARTM_WoS_Renyi_lev1}
\end{figure}

\begin{figure}
\centering
\includegraphics[width=3.3in]{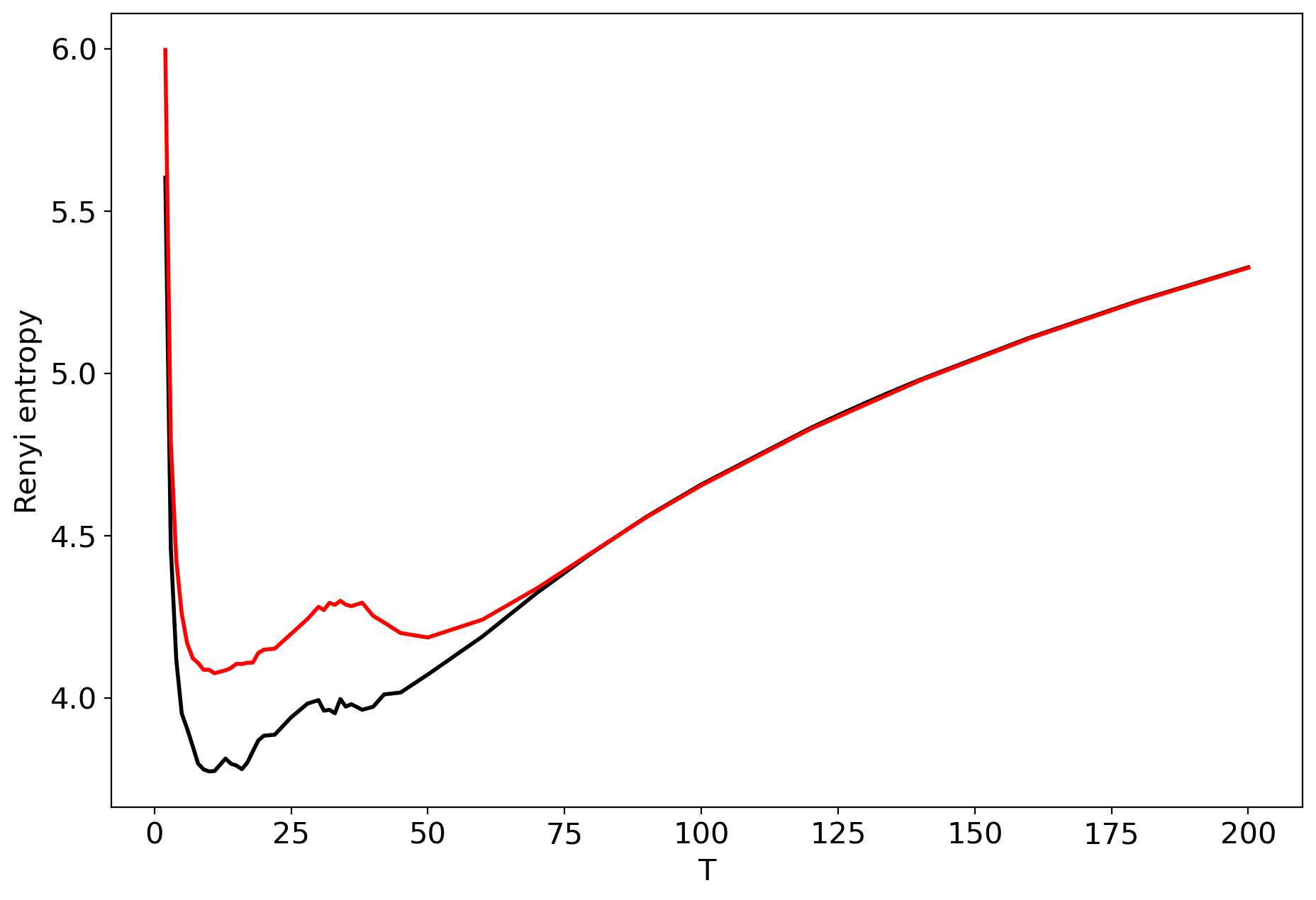}
\caption{Renyi entropy curves (hARTM). Black: balanced Amazon dataset; red: Amazon dataset.}
\label{fig:HARTM_Amazon_Renyi_lev1}
\end{figure}

Figures \ref{fig:HARTM_Lenta_Renyi_lev1}-\ref{fig:HARTM_Amazon_Renyi_lev1} demonstrate the behavior of Renyi entropy on the first level for different datasets. 
For non-hierarchical datasets, we clearly observe only one minimum of Renyi entropy. Moreover, the location of this minimum is close to the human mark-up, namely 7 topics for the Lenta dataset and 14 topics for the 20 Newsgroups dataset. The behavior of Renyi entropy on the second level is almost identical to that of the first level; therefore, we do not provide figures. Let us note that Renyi entropy curve for hARTM model does not have sharp jumps compared to the hPAM model. Thus, the entropy approach can be successfully used for determining the structure of non-hierarchical datasets. 

\begin{figure}
\centering
\includegraphics[width=3.3in]{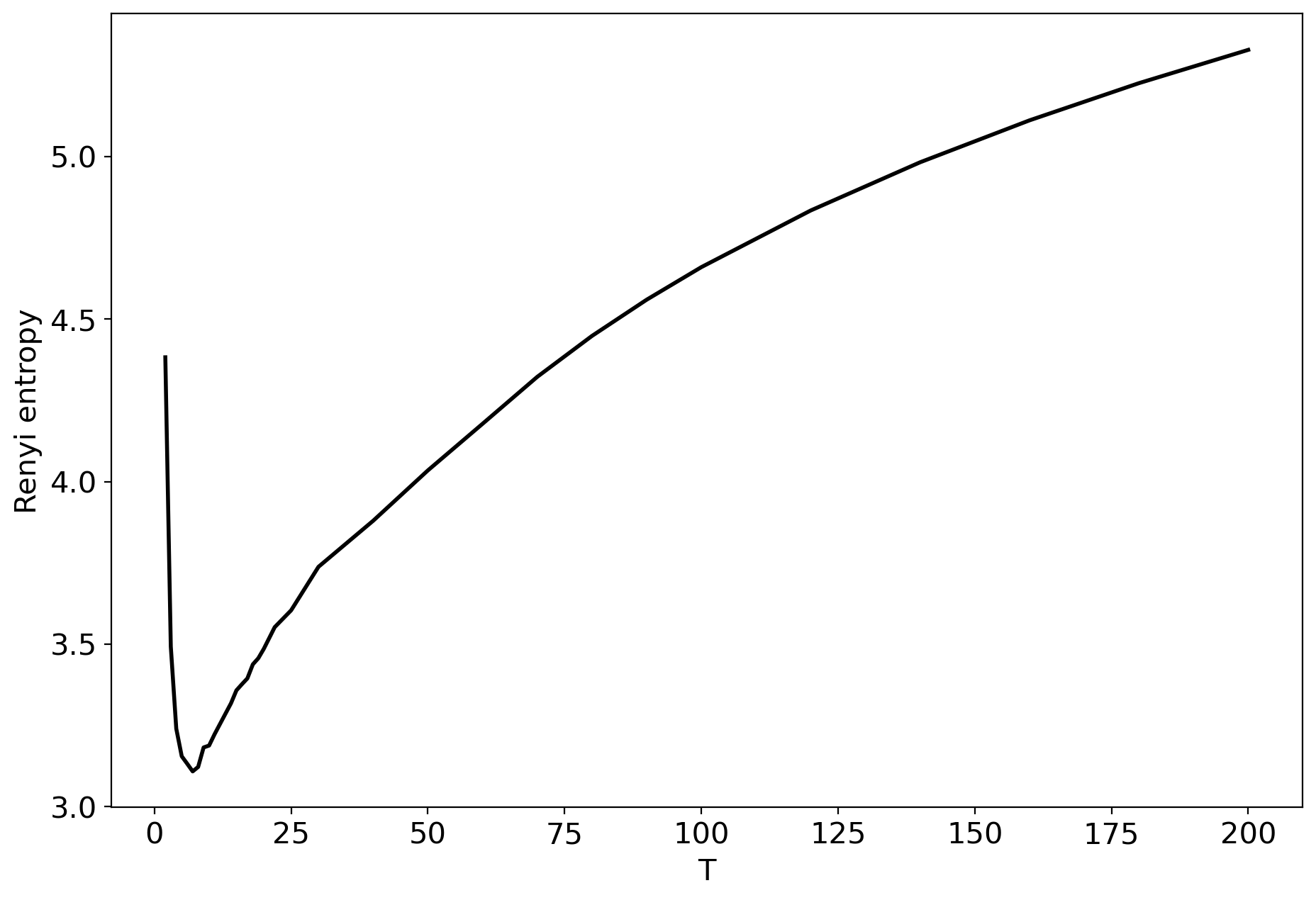}
\caption{Renyi entropy curve (hARTM). Lenta dataset.}
\label{fig:HARTM_Lenta_Renyi_lev1}
\end{figure}

For hierarchical datasets, we observe two minima of Renyi entropy that allows us to identify two levels in the data structure. The first (global) minimum of Renyi entropy corresponds to the first level of the hierarchical mark-up of the dataset and the second minimum corresponds to the second level. For instance, Renyi entropy for balanced WoS dataset has a global minimum for 10 topics, and human mark-up has 7 topics. The second (local) minimum of Renyi entropy corresponds to 36-42 topics, and the mark-up has 34 topics. For the balanced Amazon dataset, the first minimum of Renyi entropy corresponds to 10 topics, and the mark-up has 6 topics on the first level. The second minimum corresponds to 38 topics, and the mark-up has 27 topics on the second level.
Thus, the estimation of the number of topics on the second level has a larger error with respect to that of the first level. However, to the best of our knowledge, Renyi entropy approach provides the best accuracy of estimating the number of topics for this hierarchical model. Moreover, our approach allows us to determine the presence of hierarchical structure in the data. 
Let us note that balancing datasets allows us to obtain strongly pronounced topics that, in turn, improves the results of topics modeling. This leads to the occurrence of words with high probabilities that, in turn, leads to the occurrence of a more pronounced entropy minimum and decreasing of entropy on average. Figures \ref{fig:HARTM_WoS_Renyi_lev1} and \ref{fig:HARTM_Amazon_Renyi_lev1} demonstrate the difference between Renyi entropy for balanced and not balanced datasets. The effect of balancing is particularly seen in the region of entropy minima, while for large numbers of topics the effect is less observable. This is due to the fact that increasing the number of topics leads to an almost uniform distribution of word probabilities regardless of the dataset content.

\begin{figure}
\centering
\includegraphics[width=3.3in]{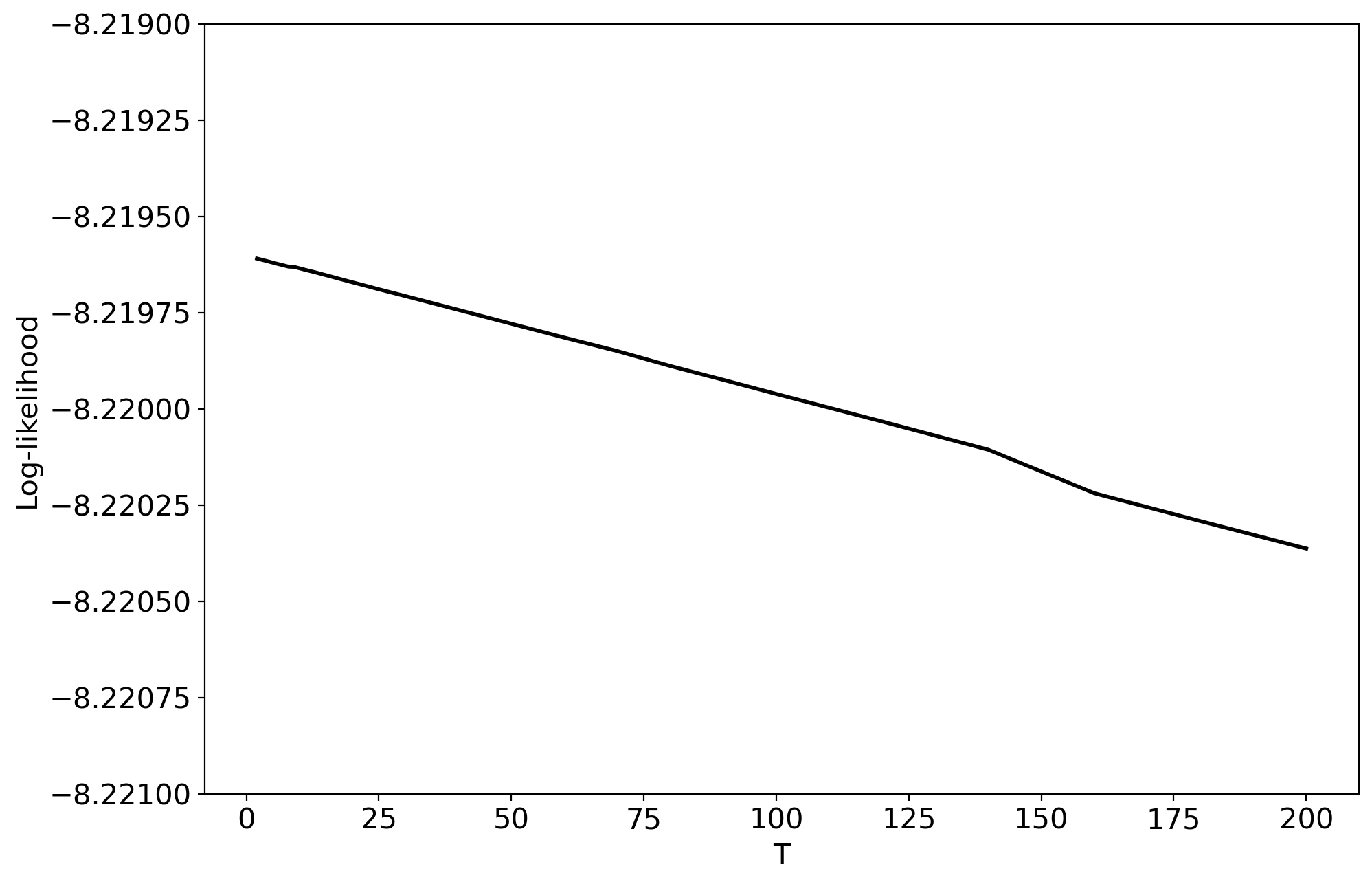}
\caption{Log-likelihood curve (hARTM). Lenta dataset.}
\label{fig:HARTM_Lenta_likelihood_lev1}
\end{figure}

\begin{figure}
\centering
\includegraphics[width=3.3in]{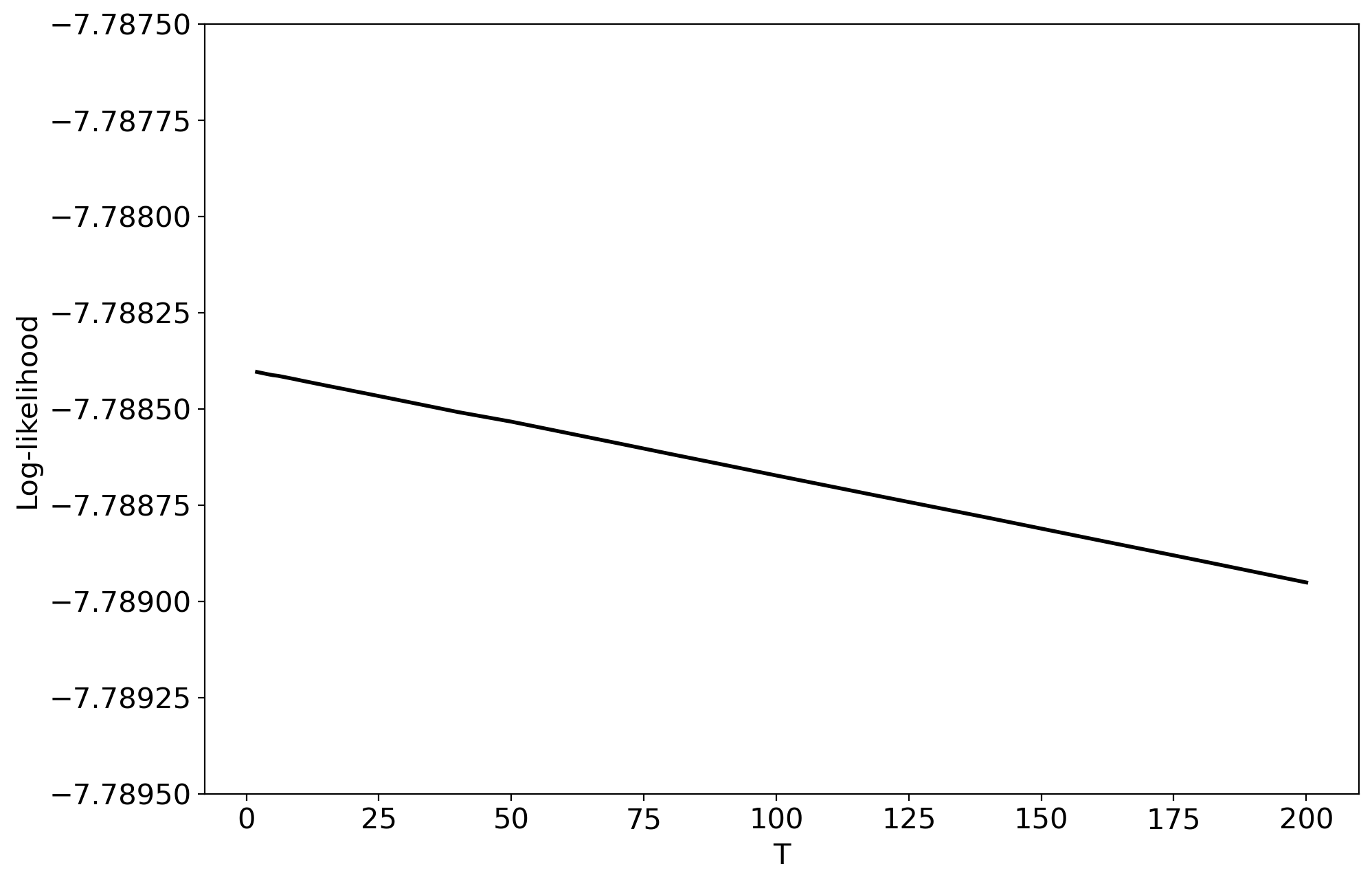}
\caption{Log-likelihood curve (hARTM). Balanced WoS dataset.}
\label{fig:HARTM_WoS_likelihood_lev1}
\end{figure}

Figures \ref{fig:HARTM_Lenta_likelihood_lev1}, \ref{fig:HARTM_WoS_likelihood_lev1} demonstrate the behavior of log-likelihood in dependence of the number of topics for non-hierarchical and hierarchical datasets. Let us note that the behavior is monotone and does not allow us to determine the dataset structure.

\section*{Discussion}
As our computer experiments demonstrate, hLDA model is very unstable which means that different runs of the model with the same parameters produce different topical structures of the same data. This instability can be significantly reduced by changing parameter $\eta$, which controls the sparsity of topics. However, how to choose the optimal value of $\eta$ is an open question since neither log-likelihood nor Renyi entropy allows us to tune this parameter.

Recall that hPAM model depends on the parameter $\eta$ and the number of topics on different levels. Variation of these parameters in computer experiments demonstrated that Renyi entropy allows us to determine only one level of a data structure for this model, namely, the number of topics on the second level. Variation of the number of topics on the next level leads to significant fluctuations of Renyi entropy for both non-hierarchical and hierarchical datasets. Correspondingly, it seems that sharp jumps of Renyi entropy on the third level are related to the model's features rather than to dataset structure. Moreover, the behavior of log-likelihood as a function of the number of topics and parameter $\eta$ does not allow us to determine the dataset structure since maxima and minima of log-likelihood do not correspond to the structure of mark-up of hierarchical and non-hierarchical datasets. 

Among the considered models, hARTM is the most stable since the procedure of initialization of this model has almost no effect on the obtained values of Renyi entropy. Moreover, Renyi entropy for this model demonstrates that for the datasets with non-hierarchical mark-up (for both English-language dataset and Russian-language dataset), there is only one minimum, which is located close to the number of topics from the mark-up. Further increasing of the number of topics does not lead to jumps (in contrast to hPAM model) but leads to smooth increasing of Renyi entropy that coincides with the results of modeling with non-hierarchical topic models (such as Latent Dirichlet allocation \citep{Blei:2003, Griffiths:2004} and granulated Latent Dirichlet Allocation \citep{Koltcov:2016}) on the same datasets \citep{Koltcov:2020}. Moreover, for datasets with hierarchical mark-up, Renyi entropy has two clear minima.  One of these minima corresponds to the mark-up of the first level, and the second corresponds to the mark-up of the second level. The difference between the number of topics obtained by searching for Renyi entropy minimum and the results of the mark-up is about 3-4 topics for the first level and 8-10 topics for the second level. Thus, estimation of the number of topics on the second level has poorer quality. However, other metrics, which are based on the estimation of the predictive power of the model, such as log-likelihood or perplexity, do not allow us to select the optimal parameters for hierarchical models. Moreover, the mark-up of text collection has a certain degree of instability; therefore, the results of determining the optimal number of topics by searching Renyi entropy minimum are significant.

The proposed Renyi entropy approach has some limitations related to the method of Renyi entropy calculation and distribution of words in text collections. First, distributions of words in text collections have long and heavy tails. Words from such a tail are assigned with small probabilities in the result of topic modeling. Since Renyi entropy measures the difference between words with high probabilities and words with small probabilities, it can only detect topics that are comprised of highly probable words. However, low levels of hierarchy may contain narrow topics, each of which is comprised of words with small probabilities. Correspondingly, the ability to distinguish the  sets of topics disappears as the number of hierarchy level increases.
Thus, Renyi entropy approach is suitable for determining the number of topics and values of hyperparameters for one or two levels in a hierarchical structure. Another limitation of our approach is the presence of fluctuations in Renyi entropy values. These fluctuations are related to fluctuations of word probabilities, which, in turn, are caused by the stochastic nature of topic modeling. The problem of minimizing the instability of topic modeling has not yet been solved, but the stabilization of topic models through word embeddings can be potentially interesting. The application of Renyi entropy approach to topic models with word embeddings is beyond the scope of this work  and could be considered as the next potential stage in the development of the entropic approach in topic modeling. 

\section*{Conclusion}
In this work, we investigated the ability of hierarchical topic models to correctly determine the hierarchical structure in data using three hierarchical topic models applied to four datasets, which have human mark-up on topics. Two of these datasets have non-hierarchical mark-up and two others have two-level mark-up. In the first part of the work, we formulated the principle of Renyi entropy calculation for hierarchical topic models. In the second part, we subsequently analyzed the chosen topic model using log-likelihood metric and Renyi entropy. Based on the obtained results of calculations, one can conclude the following. First, the application of Renyi entropy approach can be extended to hierarchical topics models since the accuracy of the approximations of the optimal number of topics for such models is not inferior to that for non-hierarchical models demonstrated in \citep{Koltsov:2018, Koltcov:2020}.  Second, calculations on the test datasets demonstrated that hLDA model is not applicable for practical tasks due to its extreme instability. Third, for hPAM model, the proposed Renyi entropy approach allows selecting the number of topics only for one hierarchical level. Determining the number of topics on the next hierarchical level is complicated due to large fluctuations of entropy for large numbers of topics. Thus, hPAM model can be used for modeling datasets with non-hierarchical topical structures. Let us note that log-likelihood metric does not allow us to tune the model parameters. Fourth, based on our calculations, one can conclude that hARTM model provides the best results since Renyi entropy approach allows us to determine both the non-hierarchical and hierarchical structure of the datasets. In the case of non-hierarchical dataset structure, we observe only one minimum of Renyi entropy under variation of the number of topics. In the case of a two-level dataset structure, Renyi entropy has two minima, each of which corresponds to one level in the hierarchy. Thus, hARTM model is the easiest for tuning and the most stable among the considered ones. The most useful metric of quality for model tuning is Renyi entropy. The limitation of Renyi entropy approach is that only one-two hierarchical levels can be determined. Subsequent hierarchical levels can not be determined since the probabilities of words on these levels are sufficiently small, and the entropic approach does not see the optimal partition for a large number of topics. The problem of small topic detection and stabilization of topic models can be possibly solved by means of word embeddings technology, however, this direction is out of scope for this work.

\section*{Acknowledgments}
This work is the result of the collaboration with Paolo Rosso in the
framework of his virtual online internship at the National Research
University Higher School of Economics due to COVID-19.\\
This research was supported in part through computational resources of HPC facilities at NRU HSE.

\section*{Funding statement}
The study was implemented in the framework of the Basic Research Program at the National Research University Higher School of Economics (HSE University) in 2020 (Project: "Online communication: cognitive limits and methods of automatic analysis").

\appendix
\section*{Appendix A}\label{appendixA}
Let us briefly discuss the most widely used topic models and the existing quality metrics in the field of topic modeling.
Assume that we have a document collection $D$ with a vocabulary of all unique words denoted by $W$. A document $d$ contains a sequence of words $\{w_1, w_2,...,w_n\}$, where $w_i\in W$. So, the words and the documents are the only observable variables. The goal of topic modeling is to retrieve hidden topics in the document collection, where each topic is characterized by its distribution over the vocabulary. Thus, the primary goal of topic modeling is to find the word distribution for each topic and to find the proportions of topics in each document. In most probabilistic topic models, it is assumed that, first, there exists a finite number $T$ of topics, and each entry of a word $w$ in document $d$ is associated with a certain topic $t$. Second, it is assumed that the order of words in documents is not important for TM (`bag-of-words' model) and, in addition, the order of documents in the collection is neither important. Third, a conditional independence assumption states that document $d$ and word $w$ are independent conditioned on the latent topic $t$, i.e., words,  conditioned on the latent class $t$, are generated independently of the specific document identity $d$ \citep{Hofmann2}. 

Two basic topic models are probabilistic latent semantic analysis (pLSA) \citep{Hofmann} and its Bayesian version called Latent Dirichlet Allocation (LDA) \citep{Blei:2003}. Mathematically, the probability of word $w$ in document $d$ can be expressed as follows  \citep{Hofmann}:

\begin{equation}\label{eq: prob}
    p(w|d)=\sum_{t\in T} p(w|t) p(t|d)=\sum_{t\in T} \phi_{wt} \theta_{td},
\end{equation}

where $t$ is a topic, $p(w|t)$ is the distribution of words by topics, and $p(t|d)$ is the distribution of topics by documents.
The output of TM is represented with two matrices, namely, matrix $\Phi:=\{\phi_{wt}\}\equiv \{p(w|t)\}$ containing the distribution of words by topics
and matrix $\Theta:=\{\theta_{td}\}\equiv \{p(t|d)\}$ containing the distribution of topics by documents (or, in other words, the proportions of topics in documents). Many modifications of the LDA model were developed for various specific applications. However, these models share the same practical issue: to build a model, the user has to set the number of topics unknown in advance in many cases.

Generally, standard quality metrics such as perplexity \citep{Heinrich:2004, Newman, Zhao}, log-likelihood \citep{Wallach, Heinrich:2004, Griffiths:2004}, and semantic coherence \citep{Mimno:2011, Stevens:2012} are used to select the number of topics or to tune other model parameters. Perplexity is a metric that evaluates the efficiency of the model to predict the data. For LDA-based models,  the perplexity for document collection $\hat{D}$ with $D$ documents can be calculated as follows \citep{Heinrich:2004}:

$$\textnormal{perplexity}(\hat{D})=\exp\left(-\frac{\sum_{d}\log p(d)}{\sum_{d} N_d}\right)= \exp\left(-\frac{\sum_{d} \sum_{w} n_d^w \ln (\sum_{t} \phi_{wt} \theta_{td})}{\sum_{d} N_d}\right),$$ 

where $N_d$ is the number of words in document $d$, $n_d^w$ is the number of times word $w$ was observed in document $d$. A lower perplexity value means better model quality. Perplexity is closely related to likelihood, namely, perplexity is a reciprocal geometric mean of the likelihood, where likelihood for a document $d$ is expressed as \citep{Heinrich:2004}  

$$P(d|\Phi, \Theta)=  \prod_{w} ( \sum_{t} \phi_{wt}\theta_{td})^{n_{d}^{w}}.$$

In turn, log-likelihood of document collection $\hat{D}$ can be calculated for LDA-based models as follows  \citep{Wallach, Heinrich:2004}:

$$\ln (P(\hat{D}|\Phi, \Theta))= \sum_{d} \sum_{w} n_{d}^{w} \ln(\sum_{t} \phi_{wt}\theta_{td}).$$

The number of topics and the other model parameters are selected when finding maximum log-likelihood~\citep{Griffiths:2004}.

Semantic coherence is another type of quality metrics. It aims at measuring the interpretability of inferred topics \citep{Mimno:2011} rather than the predictive power of the model. Semantic coherence can be calculated as an average of the topic coherence scores, where each topic coherence score is expressed as 

$$C(t, W(t))=\sum_{m=2}^{M}\sum_{l=1}^{m-1} \log (\frac{D(v_m^t, v_l^t)+1}{D(v_l^t)}),$$

$W(t)=(v_1^t,...,v_M^t)$ is a list of $M$ most probable words in topic $t$, $D(v)$ is the number of documents containing word $v$, and $D(v,v')$ is the number of documents where words $v$ and $v'$ co-occur. The typical value for $M$ is 5-20 words.

The discussion on the application of the above metrics to the task of determining the number of topics for pLSA and LDA models and limitations of these metrics can be found in papers \citep{Koltcov:2019, Koltcov:2020}.

Further development of topic models occurred in the direction of nonparametric models. 
The main idea of nonparametric topic modeling is to infer the model structure (namely, the number of topics) from the data. Theoretically, nonparametric models are able to select the number of topics automatically according to available data. Such models introduce a prior distribution on potentially infinite partitions of integers using some stochastic process that would give an advantage in the form of a higher prior probability for solutions with fewer topics.
In works \citep{HDP, Teh:2004}, a topic model based on hierarchical Dirichlet process (HDP) was first proposed. This model can be considered as an infinite extension of LDA model \citep{Heinrich:2008}. More complicated models that are based on the Indian buffet process are considered in works \citep{Chen:2012, Williamson:2010}. Detailed surveys on nonparametric models can be found in \citep{Hjort:2010, Rasmussen:2006}. However, nonparametric models possess a set of parameters that can significantly influence the inferred number of topics and results of TM in general \citep{Vorontsov:2015}. Moreover, in work \citep{Koltcov:2020}, we demonstrated that, in real applications, the number of topics inferred by HDP model does not correspond to the number of topics obtained with human judgment. Thus, the application of this type of models is complicated in practice. 

The next important step in the development of topic models was dictated by the intention to organize topics into a hierarchy. It resulted in the development of hierarchical topic models. In contrast to flat topic models (such as pLSA or LDA), the hierarchical topic models are able to discover a topical hierarchy, which comprises levels of topical abstraction. Usually, each node in the hierarchy corresponds to a topic, which, in turn, is represented by the distribution over words. Different hierarchical topic models are based on different prior assumptions on the distribution of topics and on the type of hierarchical structure to be inferred. The two most widely used in practice hierarchical topic models \citep{Liu:2016} are hierarchical latent Dirichlet allocation (hLDA) \citep{Blei:2003h} and hierarchical Pachinko allocation (hPAM) \citep{Mimno:2007}. 

hLDA model is a hierarchical and nonparametric extension of LDA model. In the framework of this model, it is assumed that each sub-topic (child topic) has a single parent topic, thus, providing a tree of topics. Moreover, it is assumed that all topics of a document are found within a path in that tree. It imposes significant restrictions on the inferred topical components of documents. Thus, for instance, a document can not be devoted to several specific sub-topics within the assumptions of the model. Another feature of this model is that the first level of the hierarchy always contains one topic (the root of the hierarchy). hLDA model learns topic hierarchies based on the nested Chinese Restaurant Process (nCRP), which specifies a prior for the model.
Nested Chinese Restaurant Process is a hierarchical version of the Chinese Restaurant process and is used to express uncertainty about possible $L$-level trees. To illustrate nCRP, assume that there are infinitely many Chinese restaurants with infinitely many tables. One restaurant is associated with a root (level 1) and each table of this restaurant has a card with a reference to another restaurant (level 2). Tables of those restaurants, in turn, have references to other restaurants (level 3), and this structure repeats.  
Each restaurant is referred to exactly once, therefore, the structure produces an infinitely-branched tree of restaurants. 
Imagine that a tourist enters the root restaurant and selects a table according to 

\begin{equation}\label{eq:CRP}
\begin{split}
&p(\textnormal{occupied table i}|\textnormal{previous customers})=\frac{m_i}{\gamma+m-1} \\
&p(\textnormal{new table}|\textnormal{previous customers})=\frac{\gamma}{\gamma+m-1},
\end{split}
\end{equation}

where $m_i$ is the number of previous customers at table $i$, $m$ is the number of customers in the restaurant, including the tourist, and $\gamma$ is a parameter that controls how often a customer chooses a new table. The next day, the tourist goes to the restaurant identified on the card of the table chosen the day before and selects a table according to \eqref{eq:CRP}. This process is repeated for $L$ days. After $M$ tourists proceed this process, the collection of their paths describes a particular $L$-level tree. In terms of topic modeling, customers correspond to documents, and restaurants correspond to topics.
In the framework of hLDA model, the following prior distributions are assumed: 1) nCRP prior with hyperparameter $\gamma$ on possible trees; 2) symmetric Dirichlet prior with hyperparameter $\eta$ on the distribution of words by topics  $\phi_{w t}$; 3) $L$-dimensional Dirichlet prior with hyperparameter $\alpha$ on mixing proportions $\theta_{td}$ of the topics along the path from the root to the leaf.

The generative process of hLDA model can be described as follows: 
\begin{itemize}
\item For each node $t$, draw $\phi_{\cdot t}\sim \textnormal{Dir}(\eta)$. 
\item For each document $d$: draw a path of topics $\textbf{c}_{d}=\{c_{d1},...,c_{dL}\}$ according to nCRP with parameter 
$\gamma$ \eqref{eq:CRP}; draw the topic mixing proportion $\theta_d \sim \textnormal{Dir}(\alpha)$. For each position $n$ of word in the document, choose the level assignment $z_{dn}\sim Mult(\theta_d)$ (as the level is chosen in the path, the topic is determined), then draw a word from the chosen topic $w_{dn}\sim Mult(\phi_{\cdot \textbf{c}_{dz_{dn}}})$.
\end{itemize}
Thus, a document is drawn by choosing an $L$-level path through the restaurants (topics) and then sampling the words from the $L$ chosen topics. 
The inference can be approximated by means of Collapsed Gibbs sampling \cite{Blei:2010, Chen:2018}. The expressions for assessment of $z_{dn}$ and $\textbf{c}_{d}$ variables are the following: 

\begin{equation}
    p(z_{dn}=l|w_{dn}=v, W, z_{\neg dn}, \textbf{c}) \propto (a_{dl}^{\neg dn}+\alpha) \frac{b_{c_{dl},v}^{\neg dn}+\eta}{s_{c_{dl}}^{\neg dn}+V \eta},
\end{equation}

\begin{equation}    
p(\textbf{c}_d=\tilde{\textbf{c}}|W, z, \textbf{c}_{\neg d}) \propto p(\tilde{\textbf{c}}|\textbf{c}_{\neg d}) \prod_{l=1}^{L} \frac{B(b_{c_l}^{\neg d}+b_{c_l}^{d}+\eta)}{B(b_{c_l}^{\neg d}+\eta)},        
\end{equation}
where $W$ is the vocabulary (set of words), $V$ is the vocabulary size, $z=\{z_{d}\}_{d=1} ^ D$, $\neg dn$ means all the tokens excluding token $w_{dn}$, $\textbf{c}=\{\textbf{c}_d\}_{d=1}^{D}$, $a_{dl}^{\neg dn}$ is a counter that equals the number of tokens in level $l$ in document $d$ excluding token $w_{dn}$, $b_{c_{dl}, v}^{\neg dn}$ is a counter that equals the number of tokens of the word $v$ assigned to topic $c_{dl}$ excluding the current token $w_{dn}$, $s_{c_{dl}}^{\neg dn}$ is a counter that equals the number of tokens assigned to topic $c_{dl}$ excluding the current token $w_{dn}$, $\tilde{\textbf{c}}=\{c_1,...,c_{L}\}$ is a path in the hierarchy,
$b_{c_l}^{\neg d}$ is a counter that equals the number of tokens assigned to topic $c_l$ excluding the tokens of document $d$, $b_{c_l}^{d}$ is the number of tokens of document $d$ that were assigned to topic $c_{l}$, $B(\cdot)$ is the multivariate beta function and $p(\tilde{\textbf{c}}|\textbf{c}_{\neg d})$ is according to nCRP (\eqref{eq:CRP}). 
The details can be found in \cite{Chen:2018}. Then, matrix $\Phi$ is calculated according to

\begin{equation}
    \phi_{wt}=\frac{c_{wt}+\eta}{\sum_{v=1}^V c_{vt}+\eta V},
\end{equation}

where a counter $c_{wt}$ equals the number of instances word $w$ was assigned to topic $t$.

We would like to note that hLDA model has three hyperparameters: 1) $\alpha$ is a parameter of Dirichlet distribution, which controls smoothing over levels in the tree; 2) $\eta$ is hyperparameter of another Dirichlet distribution, which controls the sparsity of topics; 3) $\gamma$ is a parameter of the nested Chinese restaurant process, which controls how often a document will choose new, i.e., not previously encountered paths.


In hPAM model, in contrast to hLDA, it is assumed that each child topic can be related to each node (topic) from the upper level, thus, resulting in a directed acyclic graph of topics. Therefore, in hPAM model, a child topic has several parent topics. It reduces the necessity of inferring the correct tree structure that is crucial in hLDA. Moreover, it allows the documents to contain several specific sub-topics. Moreover, hPAM model always infers a three-level hierarchy, where the first level corresponds to a root topic, the second level corresponds to super-topics, and the third level contains sub-topics.
Another important difference between hLDA and hPAM models is that hLDA is a nonparametric model, i.e., it infers the number of topics on each level automatically while hPAM model is parametric, therefore, a user has to manually select the number of topics on each level. Thus, the number of topics on each level of the hierarchy is a model parameter. Moreover, hPAM model has two hyperparameters: $\eta$ is a parameter of a prior Dirichlet distribution for $\phi_{t}$, which controls sparseness of topics, and $\alpha$ is a parameter of a prior Dirichlet distribution for $\theta$. 
Let us note that since hPAM constructs a directed acyclic graph, each node (topic) at a given level has a distribution over all nodes on the
next lower level. Let $T_1$ be the number of 'super-topics' and $T_2$ be the number of 'sub-topics'.

The generative process of hPAM model is as follows (hPAM model 2):
\begin{itemize}
\item For each vertex (topic) $t$ of the graph, draw $\phi_{\cdot t}\sim Dir(\eta)$. 
\item For each document $d$, sample $\theta_0$ from a $T_1+1$-dimensional Dirichlet distribution with hyperparameter $\alpha_0$ and distribution $\theta_{T}$ from a $T_2+1$-dimensional Dirichlet distribution with hyperparameter $\alpha_T$ for each super-topic. The first component of $\theta_0$ determines the probability of the event that a word is generated by the root topic while the other  $T_1$ components of $\theta_0$ determine the distribution of the root topic over super-topics.
Analogously, the first component of $\theta_{T}$ defines the probability of the event that a word is generated by the super-topic $T$, and the other $T_2$ components determine the distribution of the super-topic $T$ over sub-topics. 
\item For each word $w$: sample a super-topic $z_T \sim Mult(\theta_0)$. If $z_T=0$, sample a word from $\phi_{\cdot 0} \sim Dir(\eta)$. Otherwise, sample a sub-topic $z_t \sim Mult(\theta_{z_T})$. If $z_t=0$, then sample a word from $\phi_{\cdot z_T}\sim Dir(\eta)$. Otherwise, sample a word from $\phi_{\cdot z_t}$.
\end{itemize}
Thus, matrix of the distribution of words by topics has dimension $W \times (T_1+T_2+1)$. The inference algorithm is based on Gibbs sampling method \cite{Mimno:2007}. 
Sampling distribution for a given word $w$ in document $d$ is as follows:

\begin{equation}\label{eq:hpam_sample}
    p(z_{T}=x, z_{t}=y| w_{dn}=w, W, \textbf{z}_{\neg w}) \propto (a_{dx}^{\neg dn}+\alpha_{0,x}) \cdot \displaystyle\frac{b_{d,x,y}^{\neg dn}+\alpha_{x,y}}{\sum_{y=1}^{T_2}(b_{d,x,y}^{\neg dn}+\alpha_{x,y})} \cdot \displaystyle\frac{c_{wk}^{\neg dn}+\eta}{\sum_{w} (c_{wk}^{\neg dn}+\eta)},
\end{equation}

where $x\in \{0,1,..., T_1\}$ is the index of super-topic if $x\neq 0$; $y\in \{0,1,...,T_2\}$ is the index of sub-topic if $y\neq 0$; $\textbf{z}_{\neg w}$ is the topic assignments for all other words, counter $a_{dx}^{\neg dn}$  equals the number of tokens (excluding the current tokem $w_{dn}$) in document $d$ that were assigned to super-topic $x$, $x\neq 0$, and to the root topic if $x=0$; $\alpha_{0,x}$ means the $x$-th component of vector $\alpha_0$; counter $b_{d,x,y}^{\neg dn}$ equals the number of tokens (excluding the current token $w_{dn}$) in document $d$ that were assigned to super-topic $x$ and sub-topic $y$; counter $c_{wk}^{\neg dn}$ equals the number of tokens of the word $w$ that were generated by topic $k$, where $k$ is the root topic if $x=0$, $k=1,...,T_1$ is a super-topic if $y=0$ and $k=T_1+1,..., T_2+1$ is a sub-topic otherwise. 
Then, a pair of indices $(x,y)$ is sampled from the distribution \eqref{eq:hpam_sample} and assigned to the current word $w_{dn}$. 
Matrix $\Phi$ is calculated according to

\begin{equation}
    \phi_{wt}= \frac{c_{wt}+\eta}{\sum_{v=1}^V c_{vt}+\eta V},
\end{equation}

where a counter $c_{wt}$ equals the number of instances word $w$ was generated by topic $t$.
One can use the fixed point update equations described in \cite{Minka:2000} to
optimize the asymmetric Dirichlet parameters $\alpha$. Thus, in most of the publicly available model implementations hyperparameter $\eta$ is fixed and has to be set by a user while hyperparameters $\alpha_{t}$ are optimized during the training step.

Besides hLDA and hPAM, many other hierarchical topic models have been developed, for instance, hierarchical probabilistic latent semantic analysis (hPLSA) \citep{Gaussier:2002}, topic hierarchies of Hierarchical Dirichlet Processes (hHDP)  \citep{Zavitsanos:2011}, hierarchical latent tree analysis (HLTA) \citep{Chen:2017}, and hierarchical Additive Regularization of Topic Models (hARTM) \citep{Chirkova:2016}.

The generative process of hPLSA is as follows. First,  a document class $\alpha$ is chosen with probability $p(\alpha)$ from a predefined number of classes. Second, a document $d$ is chosen according to conditional probability $p(d|\alpha)$. Third, a topic for each word position is chosen according to class-conditional probability $p(t|\alpha)$. Forth, a word is sampled according to $p(w|t)$.
In the hierarchical structure of hPLSA, the document classes are the leaves of the hierarchy while topics occupy non-leaf nodes of the hierarchy. pLSA model is a special case of hPLSA since if only one topic per class is sampled then the result corresponds to the flat topic solution of pLSA.  If a topic is shared among classes, then it is placed at a higher level of the hierarchy. hPLSA model has certain limitations, which are analogous to limitations of pLSA, namely, it possesses a large number of parameters that have to be estimated, and this number grows linearly with the size of the dataset. It can lead to model overfitting.

In the framework of hHDP, two models are proposed. The first one (hvHDP) results in a hierarchy, where internal nodes are represented as probability distributions over topics and over words. Thus, in hvHDP model, all nodes can be considered as topics. In the second model (htHDP), only leaf nodes are represented as distributions over words. Thus, only the leaf nodes are essentially the topics. Both models are non-parametric and exploit the mixture model of hierarchical Dirichlet
processes (HDP). At each level of the hierarchy, there is a Dirichlet process for each document and a global Dirichlet process over all the Dirichlet processes at that level. Thus, each level is associated with a HDP. These assumptions allow  inferring the number of nodes on each level automatically. According to the authors \citep{Zavitsanos:2011}, hPAM and hLDA are the closest “relatives” of hvHDP in terms of inferred hierarchical structure. Indeed, analogously to hLDA, hHDP model is able to infer the number of topics automatically. In addition, analogously to hPAM, hHDP allows a child node to have several parent nodes, thus providing a more flexible structure. In turn, htHDP resembles the PAM model \citep{Li:2006} since the words in both models are generated only at the leaf level. In contrast to PAM, htHDP is fully non-parametric and is able to infer the depth of the hierarchy and the number of nodes at each level. However, hHDP has a set of hyperparameters ($H$, $\alpha$, $\gamma$) that can potentially lead to different hierarchies in terms of the depth and the number of nodes (topics) on each level. 

HLTA is a probabilistic hierarchical topic model, however, it has significant differences with respect to LDA-based models. First,  HLTA models a collection of documents without specifying a document generation process. The latent variables are  unobserved attributes of the documents. Second, each observed variable is related to a word and is a binary variable that represents the presence or absence of the word in a document. Third, topics in HLTA are clusters of documents. Namely, each binary latent variable in HLTA partitions a document collection into two soft clusters of documents. The document clusters are interpreted as topics. HLTA provides a tree-structured model, where the word variables are at the leaves and the latent variables are at the internal nodes. In turn, each latent variable can be described by a set of top words according to their mutual information, i.e., by a set of words that are the best ones to characterize the difference between the two clusters due to the fact that their occurrence probabilities in the two clusters differ the most. 
Latent variables at high levels of the hierarchy correspond to more general topics, while latent variables at low levels correspond to more specific topics.
The construction of the hierarchy is based on the subsequent application of flat models. The details of the model construction can be found in the original work \citep{Chen:2017}. 

hARTM is a hierarchical version of the proposed earlier Additive Regularization of Topic Models (ARTM) approach. In the framework of hARTM, it is allowed for a topic to have several parent topics and, moreover, the authors claim that the model can automatically determine the number of sub-topics for each topic \citep{Chirkova:2016}. However, the number of topics on each level of the hierarchy has to be specified by a user. To construct a hierarchy, it is proposed to learn several flat topic models and then to tie them via regularization. Thus, for already learned $\Phi^l$, i.e., the matrix containing the distribution of words by topics for topics on level $l$, it is proposed to implement matrix decomposition $\Phi^l \sim \Phi^{l+1} \Psi$, where matrix $\Psi=\{p(t^{l+1}|t^{l})\}$ contains interlevel distributions of sub-topics $t^{l+1}$ in parent topics $t^{l}$, $\Phi^{l+1}$ is the matrix containing the distribution of words by sub-topics $t^{l+1}$ with additional sparsing regularizers. So, one can infer the hierarchy level by level via finding parent topics for each sub-topic using interlevel regularizers. A more detailed description of the model and the model inference can be found in work \citep{Chirkova:2016}.

In addition to a wide variety of unsupervised hierarchical topic models, many semi-supervised and supervised extensions of hLDA have been proposed. For instance,  supervised hierarchical latent Dirichlet allocation (SHLDA) \citep{Nguyen:2013}, constrained hierarchical Latent Dirichlet Allocation (constrained-hLDA) \citep{Wang:2014}, hierarchical labeled-LDA (HLLDA) \citep{Petinot:2011}, and semi-supervised hierarchical latent Dirichlet allocation  (SSHLDA) model \citep{Mao:2012}.

Although many hierarchical topic models have been developed, there are still no well-established quality metrics for topic hierarchies \citep{Chen:2017, Zavitsanos:2011, Belyy:2018}. Therefore, hierarchical topic models are often compared by the same quality metrics as flat models, namely,  by means of per-word log-likelihood \citep{Chambers:2010, Chen:2017}, perplexity \citep{Zavitsanos:2011} or semantic coherence \citep{Chen:2017}. However, log-likelihood and perplexity are criticized for their inability to account for the interpretability of topic solutions that, in turn, is essential for end-users. Moreover, it was demonstrated that improved likelihood may lead to lower interpretability \citep{Chang:2009}. In contrast, semantic coherence is closer to the human evaluation of topic modeling output, however, it measures only topic interpretability and does not take into account the  parent-child relations in topic hierarchies \citep{Belyy:2018}. Therefore, semantic coherence only partially reflects the quality of a hierarchical model ignoring the hierarchical relations of topics. 

\section*{Appendix B}
Table \ref{tab:tab1} demonstrates minimum points of Renyi entropy on the second hierarchical level in dependence of  different values of hyperparameter $\eta$ for hPAM model for all the datasets under study.
$T_1$ refers to the number of topics on the second level of the hierarchy. Potentially interesting combinations of parameters that were used in our calculations at the second stage are highlighted in bold.

\begin{table}[ht]
\centering
\begin{tabular}{c|c|c|c}
 Dataset & $\eta$ & $T_1$ & Average minimum of $S_q^R$ \\\hline
Lenta & 0.001 & 7 &  3.539415\\
Lenta & 0.01 & 6 &  3.183375\\
Lenta & \textbf{0.2} & \textbf{6} & \textbf{3.127341} \\
Lenta & \textbf{0.3} & \textbf{8} &  \textbf{3.119127}\\
Lenta & \textbf{0.5} & \textbf{7} & \textbf{3.148767} \\
Lenta & 0.7 & 5 &  3.229694\\
Lenta & 1 & 4 &  3.382727\\
20 Newsgroups & 0.001 & 9 &  3.266364\\
20 Newsgroups & \textbf{0.01} & \textbf{10} &  \textbf{3.056406}\\
20 Newsgroups & 0.2 & 13 &  3.190124\\
20 Newsgroups & 0.3 & 10 & 3.172712 \\
20 Newsgroups & 0.5 & 9 &  3.140305\\
20 Newsgroups & \textbf{0.7} & \textbf{8} & \textbf{3.083744} \\
20 Newsgroups & 1 & 6 &  3.090616\\
Balanced WoS & 0.001 & 8 &  3.659404\\
Balanced WoS & 0.01 & 10 &  2.893595\\
Balanced WoS & 0.2 & 12 &  2.818518\\
Balanced WoS & \textbf{0.3} & \textbf{12} &  \textbf{2.797985}\\
Balanced WoS & \textbf{0.5} & \textbf{10} & \textbf{2.773128} \\
Balanced WoS & 0.7 & 8 & 2.87006 \\
Balanced WoS & 1 & 6 &  2.956012\\
WoS & 0.001 & 8 & 3.250087\\
WoS & 0.01 & 9 & 2.961168\\
WoS & \textbf{0.2} & \textbf{14} & \textbf{2.672866}\\
WoS & \textbf{0.3} & \textbf{8} & \textbf{2.682736}\\
WoS & \textbf{0.5} & \textbf{9} & \textbf{2.657161}\\
WoS & 0.7 & 5 & 2.952569\\
WoS & 1 & 6 & 3.386336\\
Balanced Amazon & 0.001 & 19 & 3.682714771\\
Balanced Amazon & 0.01 & 16 & 3.604896602\\
Balanced Amazon & 0.1 & 14 & 3.517948666\\
Balanced Amazon & 0.2 & 13 & 3.452508147\\
Balanced Amazon & \textbf{0.3} & \textbf{7} & \textbf{3.404987537}\\
Balanced Amazon & \textbf{0.4} & \textbf{7} & \textbf{3.317253256}\\
Balanced Amazon & \textbf{0.5} & \textbf{6} & \textbf{3.339819712}\\
Balanced Amazon & \textbf{0.6} & \textbf{5} & \textbf{3.32510498}\\
Balanced Amazon & 0.7 & 5 & 3.445027038\\
Balanced Amazon & 0.8 & 4 & 3.584118911\\
Balanced Amazon & 1 & 3 & 3.92866585\\
Amazon & 0.001 & 14 & 3.932673\\
Amazon & 0.01 & 15 & 3.737086\\
Amazon & 0.2 & 12 & 3.469943\\
Amazon & \textbf{0.3} & \textbf{7} & \textbf{3.399671}\\
Amazon & \textbf{0.4} & \textbf{7} & \textbf{3.340441}\\
Amazon & \textbf{0.5} & \textbf{6} & \textbf{3.293055}\\
Amazon & 0.7 & 3 & 4.006476\\
Amazon & 1 & 3 & 3.925447\\
\end{tabular}
\caption{\label{tab:tab1} Minimum points of Renyi entropy for hPAM model.}
\end{table}

\clearpage

\section*{Appendix C}
Table \ref{tab:tab2} demonstrates the results of hLDA model, namely, the range of the derived number of topics for 10 runs on each dataset and for each value of hyperparameter $\eta$.
\begin{table}[ht]
\centering
\begin{tabular}{c|c|c|c}
 Dataset & $\eta$ & $T_1$ & $T_2$ \\\hline
Lenta & 0.001 & 6-11 &  31-67\\ 
Lenta &	0.01 & 6-11 & 13-30 \\ 
Lenta &	0.2& 2-3 &5-7 \\
Lenta & 0.3& 2 & 2-4 \\
Lenta & 0.5	& 2 & 2-3 \\
Lenta & 0.7 & 2 & 2-3 \\
Lenta & 1 & 3 & 2-3 \\
20 Newsgroups& 0.001& 288-358 &911-1402\\ 
20 Newsgroups &	0.01 & 81-11 & 274-334\\ 
20 Newsgroups & 0.2	& 6-11 & 14-18\\ 
20 Newsgroups& 0.3	& 4-9 &	7-11 \\
20 Newsgroups & 0.5	& 3-5 &	5-9 \\
20 Newsgroups & 0.7 & 3-4 & 3-7 \\
20 Newsgroups & 1 &2-4 & 3-6 \\
Balanced WoS &0.001 & 482-652 & 1751- 2242\\
Balanced WoS & 0.01 & 68-93 & 325-453\\
Balanced WoS & 0.2 & 2-5 & 6-13\\
Balanced WoS & 0.3 & 2-3 &3-7\\
Balanced WoS & 0.5 & 2 &2-3\\
Balanced WoS &0.7 & 2 &2-3\\
Balanced WoS & 1 &2	& 2-3\\
Balanced Amazon & 0.001 & 108-148 &561-654\\
Balanced Amazon & 0.01 & 23-36 &108-122\\
Balanced Amazon &0.2 &3	&5-6\\
Balanced Amazon	& 0.3 & 2-3	& 3-4\\
Balanced Amazon &0.5 &2-3 &3-4\\
Balanced Amazon & 0.7 &2 &2-4\\
Balanced Amazon &1	& 2 &2-3
\end{tabular}
\caption{\label{tab:tab2} Range of the derived number of topics by hLDA model for the second ($T_1$) and the third hierarchical levels ($T_2$).}
\end{table}

\bibliography{sample}

\end{document}